\documentclass[journal]{IEEEtran} 

\IEEEoverridecommandlockouts                             

\usepackage{CJK}
\usepackage{pifont}
\usepackage{graphicx}
\usepackage{multicol}
\usepackage{booktabs}
\usepackage{amsmath,amssymb}
\usepackage{amsthm}
\usepackage[utf8]{inputenc}
\usepackage[english]{babel}
\usepackage{multirow}
\usepackage[font=small]{caption}
\usepackage{float}
\usepackage[hidelinks]{hyperref}

\usepackage{lettrine}
\usepackage[]{algorithm2e}
\usepackage{algpseudocode}
\usepackage{cite}
\usepackage{filecontents}
\usepackage{lipsum}
\usepackage{color}
\usepackage{esdiff}
\usepackage{epstopdf}
\usepackage[normalem]{ulem}
\usepackage{soul}

\newcommand{\tabincell}[2]{\begin{tabular}{@{}#1@{}}#2\end{tabular}}
\usepackage{subcaption}
\usepackage{xcolor}
\usepackage{colortbl}
\usepackage{balance}
\newcommand{\comm}[1]{}

\definecolor{pink}{rgb}{1, 0, 1}
\definecolor{orange}{rgb}{1, 0.7529, 0}
\definecolor{darkgreen}{rgb}{0, 0.8, 0}
\begin{document}
\title{Motion Planning in Dynamic Environments: A Survey from  Classical to Modern Methods}

\author{
{Zongyuan Shen$^1$}, {Yaming Ou$^2$}, {Shalabh Gupta$^3$}, {Shancheng Zhao$^1$}, {Dehua Zhou$^1$}, {Gao Wang$^1$}, \\ {Zhongqiang Ren$^4$}, {Junfeng Fan$^5$}, and {Long Cheng$^2$, \textit{Fellow, IEEE}}

\vspace{-12pt}
\thanks {$^1$College of Information Science and Technology, Jinan University, Guangzhou 510632, China.}
\thanks {$^2$School of Artificial Intelligence, University of Chinese Academy of Sciences, Beijing 100049, China.}
\thanks {$^3$Department of Electrical and Computer Engineering, University of Connecticut, Storrs, CT 06269, USA.}
\thanks {$^4$Global College, Shanghai Jiao Tong University, Shanghai 200240, China.}
\thanks {$^5$ Institute of Automation, Chinese Academy of Sciences, Beijing 100190, China.}
}

\maketitle

\thispagestyle{empty}

\begin{abstract}
Motion planning in dynamic environments requires robots to continuously adapt their paths in response to environmental changes for safe and uninterrupted navigation. While many surveys have reviewed planning in static settings, systematic reviews focused on dynamic environments remain limited. This paper presents a comprehensive survey of 138 works, primarily published between 2015 and 2025, spanning both classical and learning-based approaches. The motion planning methods are grouped into five categories based on the concepts of sampling, graph search, model predictive control, learning, and additional classical local planning approaches, including velocity obstacles, potential fields and dynamic windows. The learning techniques include supervised learning and reinforcement learning. We also discuss the role of dynamic perception in motion planning, covering techniques for detecting and modeling moving obstacles using cameras, LiDAR, and event-based sensors. The survey analyzes the principles, strengths, and limitations of each method, with particular attention to challenges unique to dynamic environments, such as prediction uncertainty, human–robot interaction, and the freezing robot problem. The survey provides researchers with a structured understanding of motion planning methods in dynamic environments.
\end{abstract}

\begin{IEEEkeywords}
Motion and path planning, dynamic environments, collision avoidance, autonomous robots.

\vspace{-8pt}

\end{IEEEkeywords}
\section{Introduction}
\begin{figure}[t]
        \centering        
        \includegraphics[width=0.4\textwidth]{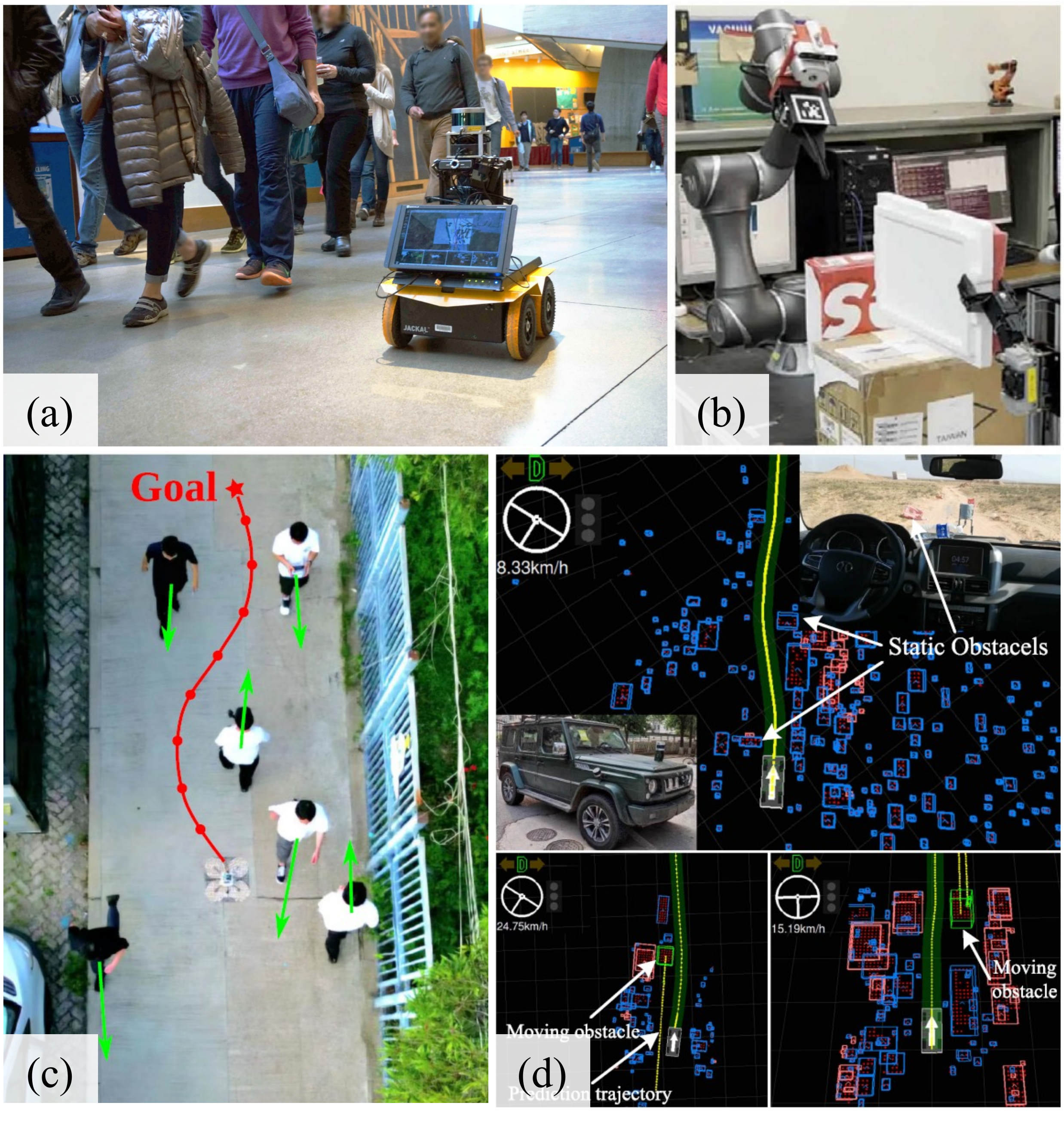}
    \caption{Examples of motion planning in dynamic environments: (a) mobile robot navigation in public spaces~\cite{Chen2017social}, (b) robotic manipulation in dynamic workspaces~\cite{lee2023path}, (c) aerial navigation among pedestrians~\cite{lu2024fapp}, and (d) autonomous driving with moving obstacles~\cite{qi2023hierarchical}.}\label{fig:example} 
    \vspace{-1em}
 \end{figure}

Motion planning in dynamic environments is a challenging problem in robotics. Recent developments in robotic platforms (e.g., autonomous vehicles) equipped with advanced sensing modalities have further accelerated the need for the development of motion planning methods for dynamic environments. Representative applications of motion planning include autonomous driving in urban traffic, multi-robot coordination in warehouses, socially aware navigation in public spaces, search and rescue operations in disaster zones, service robots in malls and hospitals, door-to-door product delivery, and human-robot collaboration in industrial environments. Fig.~\ref{fig:example} shows scenarios with moving agents such as humans, vehicles, and robots with uncertain future behaviors. Therefore, a planner must continuously adapt and generate collision-free paths for the robot in real time to accommodate environmental changes, ensure safety, and maintain uninterrupted task progression.

\begin{figure}[t]
   \centering        
    \includegraphics[width=0.50\textwidth]{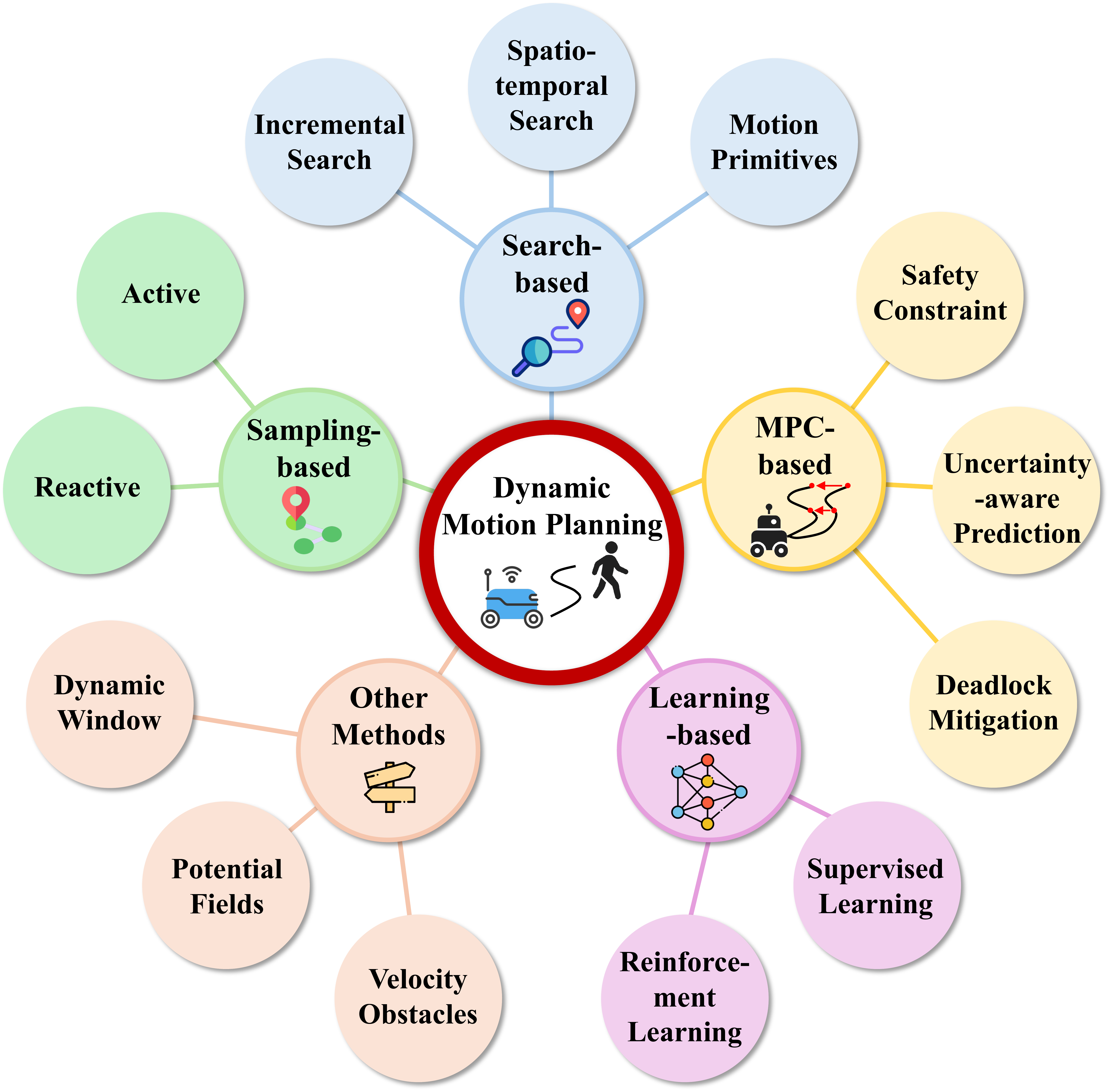} 
    \caption{Taxonomy of motion planning in dynamic environments. }\label{fig:planneroverview} 
    \vspace{-6pt}
 \end{figure}

To address the above challenges, a diverse range of methods have been developed for motion planning in dynamic environments based on the concepts of sampling, graph search, model predictive control (MPC), reinforcement learning (RL), and other approaches. Fig.~\ref{fig:planneroverview} shows the taxonomy of motion planning methods organized according to the planning mechanism. While classical methods have continued to evolve through improved replanning, risk-aware modeling, uncertainty handling, and real-time optimization, modern learning techniques are increasingly adopted either as primary planning mechanisms or as auxiliary modules for prediction, sampling guidance, cost-function tuning, and local decision-making. These developments have led to a growing number of hybrid planners that combine the interpretability and structural properties of classical methods with the adaptability of data-driven learning models. This progress necessitates a systematic and up-to-date survey that captures the diverse and fast-evolving landscape of motion planning methods in dynamic environments.  

While the literature offers several surveys on motion planning in static environments~\cite{kingston2018sampling,gammell2021asymptotically,orthey2023sampling,liu2023path}, only limited surveys exist that present comprehensive reviews specifically dedicated to motion planning in dynamic environments. Notable earlier efforts include the surveys by Kamil et al.~\cite{kamil2015review} and Mohanan et al.~\cite{mohanan2018survey}, which summarize developments up to 2015 but do not reflect the significant progress made in the past decade. More recently, Liu et al.~\cite{liu2024review} provided a focused review on motion planning for industrial manipulators operating in dynamic environments. However, an application-independent algorithm-level survey that systematically covers diverse motion planning paradigms in dynamic environments while including the supporting perception technologies is still lacking.

\begin{figure}[t]
        \centering        
        \includegraphics[width=0.40\textwidth]{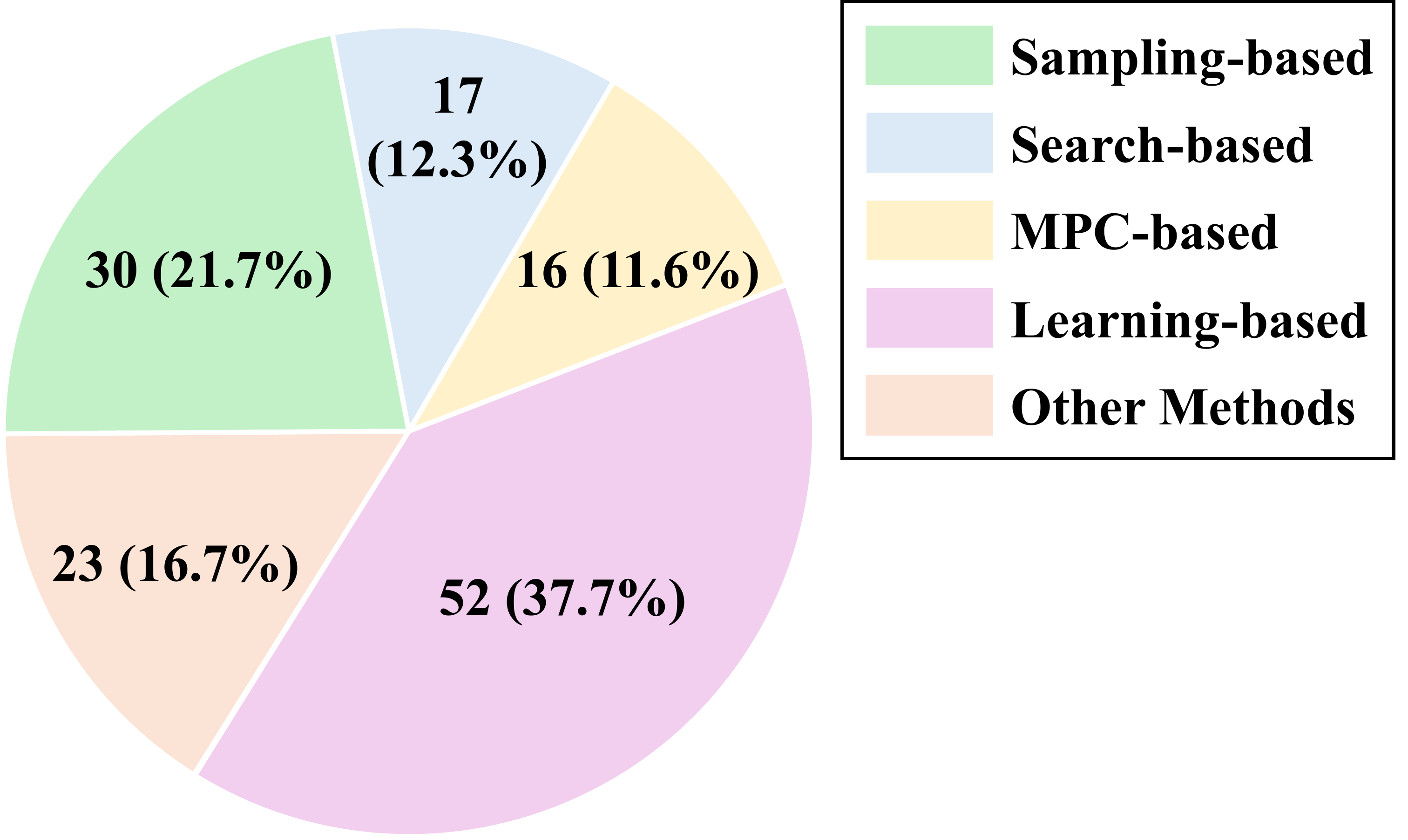}
    \caption{Category-wise distribution of motion planning methods in dynamic environments. A total of 138 references are included.}\label{fig:paperstatistics} 
    \vspace{-1em}
 \end{figure}

This survey addresses this gap by presenting a comprehensive review of motion planning methods in dynamic environments. We examined 138 representative works, primarily published between 2015 and 2025, covering a broad spectrum of approaches ranging from classical techniques to emerging learning-based and hybrid strategies. The survey highlights key challenges, traces methodological trends, and analyzes the strengths and limitations of different planning paradigms. By bridging well-established techniques with recent innovations, this survey offers a unified perspective on the current landscape of motion planning in dynamic environments. Fig.~\ref{fig:paperstatistics} shows the category-wise distribution of the reviewed works.

Furthermore, for safe and reliable motion planning in a priori unknown and dynamic scenarios, a planner depends on dynamic perception of the environment to generate situational awareness. Typically, perception is based on different sensing modalities (e.g., cameras, LiDAR, and event-based sensors), which enable robots to perceive changes in the environment, detect risks, track obstacle motions, and interpret human interactions. Thus, while the central focus of this survey is on motion planning, it includes a discussion of dynamic perception as a supporting component to provide completeness.

 \begin{figure*}[t]
        \centering        \includegraphics[width=0.95\textwidth]{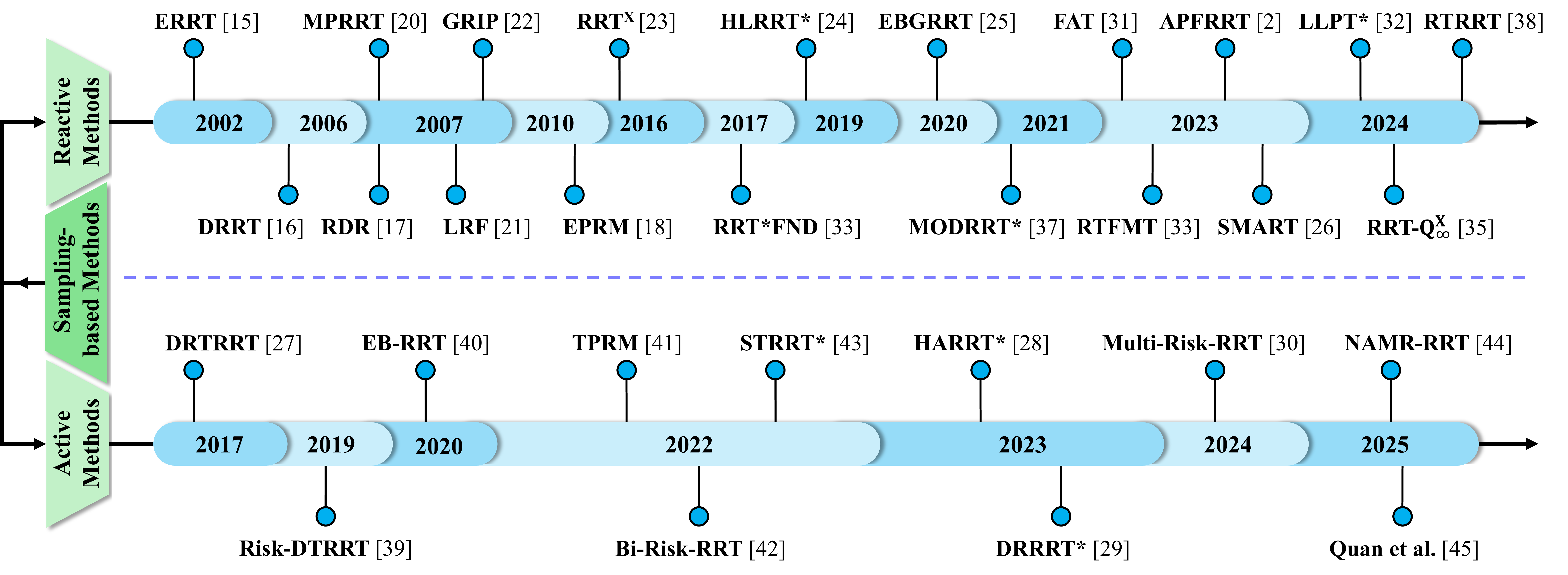}
    \caption{Timeline of the evolution of sampling-based motion planning methods in dynamic environments.}\label{fig:samplingbasedTimeMap} 
    \vspace{-1.0em}
 \end{figure*}

The remainder of this paper is organized as follows. Section~\ref{sec:samplingbasedmethod} reviews sampling-based methods. Section~\ref{searchbasedmethod} covers search-based approaches. Section~\ref{mpc} discusses model predictive control techniques. Section~\ref{learnmethod} explores learning-based methods, including supervised learning and reinforcement learning. Section~\ref{other} presents additional classical local planning strategies, including velocity obstacle-based, potential field-based, and dynamic window approaches. Section~\ref{perception} introduces dynamic perception methods. Finally, Section~\ref{conclusions} concludes the paper along with future directions.
\vspace{-3pt}
\section{Sampling-based Methods}
\label{sec:samplingbasedmethod}

The sampling-based methods developed for dynamic environments are extensions of the classical sampling-based planners originally designed for static environments, such as Probabilistic Roadmap (PRM)~\cite{kavraki1996probabilistic}, Rapidly-exploring Random Tree (RRT)~\cite{lavalle2001randomized}, and optimal Rapidly-exploring Random Tree (RRT*)~\cite{karaman2011sampling}. These methods are broadly categorized into reactive and active approaches, depending on whether the predictive information about dynamic obstacles is available. Reactive methods rely solely on the current observations of the environment to adapt the path on the fly. Although they provide fast responses, they lack foresight, which can sometimes compromise safety and reliability in complex scenarios. In contrast, active methods predict obstacle trajectories to proactively replan the paths that are safe and provide formal guaranties. However, the performance of active methods can degrade in highly dynamic or densely populated environments, where accurate prediction becomes difficult due to limited sensor coverage, high obstacle density, and computational constraints. Fig.~\ref{fig:samplingbasedTimeMap} shows the timeline of major developments in sampling-based motion planning. Table~\ref{tab:samplingbasedfeature} summarizes the key features of representative sampling-based methods.

\vspace{-6pt}
\subsection{Reactive Methods}
Reactive motion planning methods constantly monitor environmental changes and rapidly adjust the robot's path to ensure safety and task progression. These methods emphasize fast replanning through a variety of pruning, sampling, and tree-regrowing strategies, making them particularly effective in unknown or highly dynamic environments with unpredictable obstacle behavior. Instead of replanning from scratch, reactive planners typically update or repair existing paths, which significantly reduces computational overhead. 

Bruce and Veloso~\cite{bruce2002real} developed Extended RRT (ERRT), which discards the entire tree when the current path is obstructed by dynamic obstacles. A new tree is then initialized at the robot's current state and expanded by biasing samples toward the previous path to accelerate replanning process. To enhance efficiency, Ferguson et al.~\cite{ferguson2006replanning} introduced Dynamic RRT (DRRT), which prunes only the invalidated nodes and their successors, thus preserving most of the original tree structure. DRRT also biases sampling toward the trimmed region and roots the tree at the goal, ensuring consistency in the tree structure as the robot moves. Gayle et al.\cite{gayle2007reactive} developed Reactive Deforming Roadmap (RDR), which adjusts the positions of affected nodes and deforms connecting edges to preserve path feasibility. Similarly, Yang and Brock\cite{yang2010} proposed the Elastic Probabilistic Roadmap (EPRM), which uses the Elastic Band method~\cite{quinlan1993elastic} to adjust the roadmap. 

Zucker et al.\cite{zucker2007multipartite} presented Multipartite RRT (MPRRT), which prunes the colliding nodes to form multiple disconnected trees. These disjoint trees are then reconnected to the main robot-rooted tree using a forest biasing strategy. Gayle et al.\cite{gayle2007lazy} proposed Lazy Reconfiguration Forest (LRF), which maintains multiple disconnected trees after pruning invalid nodes and attempts to reconnect nearby trees through random sampling. Bekris and Kavraki~\cite{bekris2007greedy} presented a method called GRIP, which prunes disconnected branches and biases new tree growth toward the trimmed regions.

\begin{table*}[t]{}
\footnotesize
\caption {Qualitative comparison of representative sampling-based motion planning methods in dynamic environments.}\label{tab:samplingbasedfeature}\vspace{-3pt}
\centering
\setlength\tabcolsep{3pt}
\begin{tabular}{l l l l l l l l l} 
 \toprule
\specialrule{0.1em}{1pt}{1pt} 
\tabincell{l}{\textbf{Category}} 
&\tabincell{l}{\textbf{Method}}
&\tabincell{l}{\textbf{Paper}}
&\tabincell{l}{\textbf{Year}}
&\tabincell{l}{\textbf{Base}} 
&\tabincell{l}{\textbf{Pruning Portion}}  
&\tabincell{l}{\textbf{Post-pruning Structure}}
&\tabincell{l}{\textbf{Key Sampler}}
&\tabincell{l}{\textbf{Replanning Strategy}}\\ 
\toprule

% \specialrule{0em}{1pt}{0pt}
\multirow{20}{*}{\textbf{Reactive}} 
& \tabincell{l}{ERRT}
&\tabincell{l}{\cite{bruce2002real}}
&\tabincell{l}{2002}
& \tabincell{l}{{RRT}}
& \tabincell{l}{{Entire tree}}
& \tabincell{l}{None}
& \tabincell{l}{{Old path biasing}}
& \tabincell{l}{Grows a new tree from scratch}\\

\specialrule{0em}{2pt}{2pt}
\multirow{22}{*}{} 
&\tabincell{l}{DRRT}
&\tabincell{l}{\cite{ferguson2006replanning}}
&\tabincell{l}{2006}
& \tabincell{l}{{RRT}}
& \tabincell{l}{{Colliding nodes}\\{and successors}}
& \tabincell{l}{Single tree}
& \tabincell{l}{Trimmed area biasing}
& \tabincell{l}{{Grows the tree by biasing}\\{samples to the trimmed area}}\\

\specialrule{0em}{2pt}{2pt}
\multirow{22}{*}{} 
&\tabincell{l}{MPRRT}
&\tabincell{l}{\cite{zucker2007multipartite}}
&\tabincell{l}{2007}
& \tabincell{l}{{RRT}}
& \tabincell{l}{{Colliding nodes}}
& \tabincell{l}{Multiple trees}
& \tabincell{l}{Forest biasing}
& \tabincell{l}{{Reconnects the robot-rooted tree}\\{to other trees via new samples}}\\

\specialrule{0em}{2pt}{2pt}
\multirow{22}{*}{} 
&\tabincell{l}{EPRM}
&\tabincell{l}{\cite{yang2010}}
&\tabincell{l}{2010}
& \tabincell{l}{{PRM}}
& \tabincell{l}{None}
& \tabincell{l}{Graph}
& \tabincell{l}{None}
& \tabincell{l}{{Relocates the colliding}\\{nodes and deforms the edges}}\\

\specialrule{0em}{2pt}{2pt}
\multirow{22}{*}{} 
&\tabincell{l}{RRT$^\text{X}$}
&\tabincell{l}{\cite{otte2016rrtx}}
&\tabincell{l}{2016}
& \tabincell{l}{RRT*}
& \tabincell{l}{None}
& \tabincell{l}{{Graph}}
& \tabincell{l}{Random sampler}
& \tabincell{l}{{Repairs the goal-rooted}\\{subtree by rewiring cascade}}\\

\specialrule{0em}{2pt}{2pt}
\multirow{22}{*}{} 
&\tabincell{l}{HLRRT*}
&\tabincell{l}{\cite{chen2019horizon}}
&\tabincell{l}{2019}
& \tabincell{l}{RRT*}
& \tabincell{l}{{Colliding nodes}\\{and successors}}
& \tabincell{l}{Single tree}
& \tabincell{l}{GMM biasing}
& \tabincell{l}{{Biases the tree growth to}\\{promising area captured by GMM}}\\

\specialrule{0em}{2pt}{2pt}
\multirow{22}{*}{} 
&\tabincell{l}{EBGRRT}
&\tabincell{l}{\cite{yuan2020efficient}}
&\tabincell{l}{2020}
& \tabincell{l}{RRT}
& \tabincell{l}{{Colliding path nodes}\\{and successors}}
& \tabincell{l}{Two trees}
& \tabincell{l}{Old path biasing}
& \tabincell{l}{{Grows the robot-rooted tree}\\{towards goal tree for reconnection}}\\

\specialrule{0em}{2pt}{2pt}
\multirow{22}{*}{} 
&\tabincell{l}{APFRRT}
&\tabincell{l}{\cite{lee2023path}}
&\tabincell{l}{2023}
& \tabincell{l}{{RRT}}
& \tabincell{l}{{Colliding nodes}\\{and successors}}
& \tabincell{l}{Single tree}
& \tabincell{l}{APF biasing}
& \tabincell{l}{{Uses artificial potential field to}\\{guide the tree growth}}\\

\specialrule{0em}{2pt}{2pt}
\multirow{22}{*}{} 
&\tabincell{l}{SMART}
&\tabincell{l}{\cite{shen2023smart}}
&\tabincell{l}{2023}
& \tabincell{l}{RRT*}
& \tabincell{l}{{Colliding nodes}}
& \tabincell{l}{Multiple trees}
& \tabincell{l}{Informed sampling}
& \tabincell{l}{{Reconnects trees at hot-spots}\\{in an informed manner}}\\

\specialrule{0.05em}{2pt}{2pt}
\multirow{8}{*}{\textbf{Active}} 
&\tabincell{l}{DRTRRT}
&\tabincell{l}{\cite{chiang2017dynamic}}
&\tabincell{l}{2017}
& \tabincell{l}{RRT}
& \tabincell{l}{{High-risk nodes}}
& \tabincell{l}{Single tree}
& \tabincell{l}{Random sampler}
& \tabincell{l}{{Grows the tree via new samples}\\{filtered by adaptive risk tolerance}}\\

\specialrule{0em}{2pt}{2pt}
\tabincell{c}{}
&\tabincell{l}{HARRT*}
&\tabincell{l}{\cite{cai2023human}}
&\tabincell{l}{2023}
& \tabincell{l}{{RRT*}}
& \tabincell{l}{{High-risk nodes}}
& \tabincell{l}{Single tree}
& \tabincell{l}{Random sampler}
& \tabincell{l}{{Grows the tree by adding new}\\{nodes with low risk}}\\

\specialrule{0em}{2pt}{2pt}
\tabincell{c}{}
&\tabincell{l}{DRRRT*}
&\tabincell{l}{\cite{hakobyan2023distributionally}}
&\tabincell{l}{2023}
& \tabincell{l}{{RRT*}}
& \tabincell{l}{None}
& \tabincell{l}{Two trees}
& \tabincell{l}{Random sampler}
& \tabincell{l}{{Grows the full tree and updates}\\{its safe subtree to find a path}}\\

\specialrule{0em}{2pt}{2pt}
\tabincell{c}{}
&\tabincell{l}{{Multi-Risk}\\{-RRT}}
&\tabincell{l}{\cite{sun2024multi}}
&\tabincell{l}{2024}
& \tabincell{l}{{RRT}}
& \tabincell{l}{High-risk nodes}
& \tabincell{l}{Multiple trees}
& \tabincell{l}{Random sampler}
& \tabincell{l}{{Grows multiple trees in parallel}\\{to guide main-tree expansion}}\\

\bottomrule
\specialrule{0.1em}{1pt}{1pt}
\end{tabular}
\vspace{-0.5em}
\end{table*}

To fully reuse previous search efforts, Otte and Frazzoli~\cite{otte2016rrtx} proposed the RRT$^\text{X}$ algorithm, which maintains and refines the same graph without pruning any nodes. This is because previously colliding nodes may become valid as obstacles move in dynamic environments. Upon environmental changes, RRT$^\text{X}$ employs RRT*-like rewiring cascades to remodel the graph and repair the goal-rooted subtree, ensuring an updated shortest feasible path. Notably, RRT$^\text{X}$ is the first sampling-based planner for dynamic environments that is both asymptotically optimal and single-query. Several subsequent works have extended RRT$^\text{X}$ to improve performance and broaden its applicability. Huang and Jing~\cite{huang2023fast,huang2024asymptotically} enhanced RRT$^\text{X}$ by (i) accelerating convergence through selective expansion of promising vertices, and (ii) reducing computation by heuristically evaluating edge quality before computing exact costs. Silveira et al.~\cite{silveira2023real} introduced the Real-Time Fast Marching Tree (RT-FMT) algorithm, which repairs the tree via RRT$^\text{X}$-style rewiring cascades and then expands it using the FMT* algorithm~\cite{janson2015fast} to discover new paths. Xu et al.~\cite{Xu2024} extended RRT$^\text{X}$ to RRT-Q$^\text{X}_\infty$ targeting kinodynamic motion planning with unknown dynamics and disturbances. This method leverages RRT$^\text{X}$ to generate waypoints and employs a Q-learning-based controller for local waypoint navigation.

Chen et al.~\cite{chen2019horizon} proposed the Horizon-based Lazy RRT* (HLRRT*), which restricts feasibility checks to a finite time horizon, as distant dynamic obstacles pose no immediate risk to the robot. This lazy feasibility checking strategy delays expensive computations. Infeasible nodes and their successors are pruned, and the remaining tree is expanded by biasing samples toward regions of low heuristic total cost. Such regions are captured by a Gaussian Mixture Model trained online. Building on~\cite{adiyatov2017novel}, Yuan et al.\cite{yuan2020efficient} proposed Efficient Bias-goal Factor RRT (EBGRRT), which maintains both a robot-rooted main tree and a goal-rooted tree. Colliding nodes are pruned, and the main tree is guided toward the goal tree by biasing samples to the previous path nodes. 

Qi et al.\cite{qi2021} introduced Multi-objective Dynamic RRT* (MODRRT*), which prunes infeasible nodes to form multiple disjoint trees and then connects these trees directly to the nearest nodes in the goal-rooted tree. Lee et al.\cite{lee2023path} developed Artificial Potential Field RRT (APFRRT), which uses repulsive and attractive forces for obstacle avoidance and goal-seeking, respectively. Cui et al.~\cite{cui2024rt} proposed Reverse Tree-guided RRT (RT-RRT), which builds a goal-rooted reverse tree offline to generate an initial path. During execution, when the path becomes infeasible, a forward tree is grown from the robot and reconnected with the reverse tree to generate a new path.

Many of the aforementioned algorithms rely on random and biased sampling strategies, which can result in slow tree growth in cluttered environments. To address this limitation, Shen et al.\cite{shen2023smart} proposed Self-Morphing Adaptive Replanning Tree (SMART), which performs fast tree-repair at hot-spots while allowing for random sampling if necessary. Hot-spots are regions that lie at the intersection of different disjoint trees created after pruning of invalid tree portions.  The quality of each hot-spot is assessed by its utility, which is computed using the shortest-path heuristics. The algorithm incrementally selects hot-spots according to their utility values, merging disjoint trees until a new feasible path is discovered. By utilizing a maximum tree structure and quick repairs at hot spots, this method provides high computational efficiency for real-time implementation in real-world dynamic environments.

\vspace{-6pt}

\subsection{Active Methods} Active motion planning methods typically estimate the collision risk at each node by predicting the future behavior of dynamic obstacles. The risk information is then used to guide risk-aware tree pruning and expansion, enabling the planner to find safer paths within a predictive time horizon. 

Chiang et al.~\cite{chiang2017dynamic} proposed Dynamic Risk Tolerance RRT (DRTRRT), which gradually increases risk tolerance when adding new nodes over time. This design accounts for the growing uncertainty in long-term predictions, allowing the planner to avoid imminent collisions while still generating a feasible long-term path. This strategy is particularly suitable for navigation in crowded environments. Chi et al.~\cite{chi2019risk} proposed Risk-based Dual-Tree RRT (Risk-DTRRT), which first constructs a time-embedded tree by expanding low-risk nodes up to a predictive time horizon, followed by tree optimization through a rewiring process to improve path quality. Wang et al.~\cite{wang2020eb} proposed Elastic Band-based RRT (EB-RRT), which grows a time-embedded tree to generate an initial path that is further optimized by Elastic Band method~\cite{quinlan1993elastic}. Huppi et al.~\cite{huppi2022t} proposed Temporal Probabilistic Roadmap (TRPM), which predicts the future trajectories of dynamic obstacles using a constant velocity model. The node risk is modeled as binary-valued, indicating collision or not. Each node is associated with a time availability interval, during which it remains collision-free. The A* algorithm is then applied to find a globally feasible path, accounting for both the temporal constraints and time monotonicity during planning. 

Ma et al.~\cite{ma2022bi} proposed  Bidirectional Risk-RRT (Bi-Risk-RRT), which grows a forward tree rooted at the robot and a reverse tree rooted at the goal. Once the two trees connect, a path is extracted from the reverse tree to define a heuristic sampling region, which guides the forward tree toward the goal more efficiently. Grothe et al.~\cite{grothe2022st} developed the Space-Time RRT* (ST-RRT*), which constructs a forward tree from the start and a set of reverse trees from the goal in the space-time domain. The time horizon is incrementally expanded until a feasible solution is found. Cai et al.~\cite{cai2023human} proposed Human-Aware RRT* (HARRT*), which expands the tree by adding low-risk nodes up to a predictive horizon and then extracts a partial path that minimizes a composite cost combining collision probability, heuristic distance to goal, and the risk of entering dense human crowds. 

Hakobyan and Yang~\cite{hakobyan2023distributionally} proposed Distributionally Robust RRT* (DRRRT*), where node risk is quantified as the Conditional Value-at-Risk (CVaR) that represents the expected safety loss exceeding a certain risk threshold. DRRRT* maintains two structures during navigation: a full tree with all nodes and a safe subtree that contains only low-risk nodes within the predicted time horizon. Unsafe nodes can be added to the safe subtree later once their predicted risk decreases. Sun et al.~\cite{sun2024multi} proposed Multi-directional Risk-based Rapidly-exploring Random Tree (Multi-Risk-RRT), which deploys multiple trees to explore the environment in parallel. The robot-rooted main tree is guided by heuristic information provided by the other trees, enabling faster and safer progression toward the goal. This approach is later extended in Neural Adaptive Multi-Risk-RRT (NAMR-RRT)~\cite{Sun2025}, which incorporates neural network-generated heuristic regions to guide the exploration process. Quan et al.~\cite{quan2025state} constructed a State-Time Space based on the Euclidean Signed Distance Field to represent predicted pedestrian occupancy, searched for a local collision-free path using RRT*, and refined the resulting path through spatiotemporal polynomial optimization of waypoints and segment durations.

\vspace{-6pt}

\subsection{Summary of Sampling-based Methods}
Sampling-based planning methods have evolved from rebuilding trees or roadmaps to address environmental changes to computationally efficient methods that repair and utilize existing search structures. Reactive methods mainly improve replanning efficiency by preserving valid portions of trees or roadmaps, reconnecting disconnected components, or biasing samples toward informative regions. Active methods further incorporate temporal and risk information to improve safety within a finite prediction horizon. 

Despite these advances, several trade-offs remain. Although, reuse of existing search structures reduces computation, it may retain suboptimal or poorly distributed samples. On the other hand, risk-aware expansion improves safety, but it depends on prediction accuracy and may become overly conservative under uncertainties. 
Therefore, the central challenge for sampling-based planning in dynamic environments is to balance real-time responsiveness and path quality.

\begin{table*}[t]{}
\footnotesize
\caption {Qualitative comparison of representative search-based motion planning methods in dynamic environments.}\label{tab:searchbasedmethod}\vspace{-3pt}
\centering
\setlength\tabcolsep{4.5pt}
\begin{tabular}{l l l l l l} 
 \toprule
\specialrule{0.1em}{1pt}{1pt} 
\tabincell{l}{\textbf{Category}} 
&\tabincell{l}{\textbf{Method}}
&\tabincell{l}{\textbf{Paper}}
&\tabincell{l}{\textbf{Year}}
&\tabincell{l}{\textbf{Search Representation}} 
&\tabincell{l}{\textbf{Method Summary}}\\ 
\toprule

% \specialrule{0em}{1pt}{0pt}
\multirow{5.8}{*}{\textbf{Incremental Search}} 
& \tabincell{l}{D* Lite}
&\tabincell{l}{\cite{koenig2002d}}
&\tabincell{l}{2002}
& \multirow{6}{*}{\tabincell{l}{{Predefined spatial graph}\\{with dynamically updated}\\{edge costs}}}
& \tabincell{l}{{Repairs previous search results by propagating}\\{edge-cost changes through inconsistent vertices}}\\

\specialrule{0em}{2pt}{2pt}
\multirow{5.8}{*}{} 
&\tabincell{l}{MOPBD*}
&\tabincell{l}{\cite{ren2022multi}}
&\tabincell{l}{2022}
&
& \tabincell{l}{{Removes affected path labels and regenerates}\\{inconsistent states to update the Pareto-optimal path set}}\\

\specialrule{0em}{2pt}{2pt}
\multirow{5.8}{*}{} 
&\tabincell{l}{BLPA*}
&\tabincell{l}{\cite{li2023bidirectional}}
&\tabincell{l}{2023}
&
& \tabincell{l}{{Combines forward and backward searches to reuse previous}\\{search results and limits the forward search to the sensing range}}\\

\specialrule{0.05em}{2pt}{2pt}
\multirow{8.8}{*}{\textbf{Spatiotemporal Search}} 
&\tabincell{l}{SIPP}
&\tabincell{l}{\cite{phillips2011sipp}}
&\tabincell{l}{2011}
& \multirow{8.8}{*}{\tabincell{l}{{Predefined Spatiotemporal}\\{graph with dynamically}\\{updated edge costs}}}
& \tabincell{l}{{Computes safe intervals from predicted obstacle}\\{trajectories and searches over configuration-interval states}}\\

\specialrule{0em}{2pt}{2pt}
\tabincell{c}{}
&\tabincell{l}{MO-SIPP}
&\tabincell{l}{\cite{ren2022MO}}
&\tabincell{l}{2022}
&
& \tabincell{l}{{Combines SIPP safe intervals with multi-objective}\\{label search to compute Pareto-optimal collision-free paths}}\\

\specialrule{0em}{2pt}{2pt}
\tabincell{c}{}
&\tabincell{l}{Cao et al.}
&\tabincell{l}{\cite{cao2019dynamic}}
&\tabincell{l}{2019}
&
& \tabincell{l}{{Builds a Delaunay-triangulation graph from pedestrian}\\{positions and searches a collision-free path on it}}\\

\specialrule{0em}{2pt}{2pt}
\tabincell{c}{}
&\tabincell{l}{Huang et al.}
&\tabincell{l}{\cite{huang2025safe}}
&\tabincell{l}{2025}
&
& \tabincell{l}{{Constructs a safe-interval visibility graph, extracts spatial-}\\{temporally distinct paths, and optimizes B-spline trajectories}\\{within spatial-temporal corridors}}\\

\specialrule{0.05em}{2pt}{2pt}
\multirow{14.8}{*}{\textbf{Motion Primitive}} 
&\tabincell{l}{Li et al.}
&\tabincell{l}{\cite{lin2021search}}
&\tabincell{l}{2021}
& \multirow{14.8}{*}{\tabincell{l}{{Incrementally constructed}\\{spatial graph based on}\\{motion primitives}}}
& \tabincell{l}{{Generates motion primitives by discretizing control inputs,}\\{improves efficiency via primitive aggregation and pruning, and}\\{accelerates collision checking by evaluating point-line distance}}\\

\specialrule{0em}{2pt}{2pt}
\tabincell{c}{}
&\tabincell{l}{RAST}
&\tabincell{l}{\cite{chen2022rast}}
&\tabincell{l}{2022}
& 
& \tabincell{l}{{Predicts future occupancy using a particle-based dynamic map,}\\{evaluates collision risk along sampled motion primitives, and}\\{builds spatiotemporal safety corridors for trajectory optimization}}\\

\specialrule{0em}{2pt}{2pt}
\tabincell{c}{}
&\tabincell{l}{RAS}
&\tabincell{l}{\cite{chen2023risk}}
&\tabincell{l}{2023}
&
& \tabincell{l}{{Generates local risk-aware trajectories using two-phase motion}\\{primitives for collision avoidance and merges back to the global}\\{reference path when it turns safe}}\\

\specialrule{0em}{2pt}{2pt}
\tabincell{c}{}
&\tabincell{l}{Qi et al.}
&\tabincell{l}{\cite{qi2023hierarchical}}
&\tabincell{l}{2023}
&
& \tabincell{l}{{Obtains a coarse path using spatiotemporal motion primitives,}\\{combined heuristics, and fast collision checking, then refines}\\{the path and speed profile through multilayer optimization}}\\

\specialrule{0em}{2pt}{2pt}
\tabincell{c}{}
&\tabincell{l}{SLP}
&\tabincell{l}{\cite{wiman2025safe}}
&\tabincell{l}{2025}
&
& \tabincell{l}{{Uses high-resolution short-horizon lattice planning and low-}\\{resolution long-horizon guidance, switching to safe partial}\\{trajectories when no complete path to the goal is available}}\\

\bottomrule
\specialrule{0.1em}{1pt}{1pt}
\end{tabular}
\vspace{-1.0em}
\end{table*}

\section{Search-based Methods}
\label{searchbasedmethod}

Search-based methods have been extensively studied for robot navigation in static environments. These methods typically represent the configuration space as a graph or a discretized state lattice and use heuristic-guided search to generate feasible paths. However, in dynamic environments where obstacle poses change frequently, traditional graph search algorithms (e.g., Dijkstra's~\cite{dijkstra1959note} and A*~\cite{hart1968formal}) become inefficient, as they are not designed for real-time updates. To address this limitation, recent search-based methods use incremental searching, spatiotemporal reasoning, motion primitives, and hierarchical planning strategies. Table~\ref{tab:searchbasedmethod} provides a summary of representative search-based methods.

\vspace{-3pt}
\subsection{Incremental Search-based Methods}

Traditional methods, including D*~\cite{stentz1995focussed}, D* Lite~\cite{koenig2002d}, and Lifelong Planning A* (LPA*)~\cite{koenig2004lifelong}, incrementally repair A*-like solutions when edge costs are updated due to newly acquired environmental information. These methods avoid full replanning by updating only the affected parts of the graph, thereby improving computational efficiency. 

Ren et al.~\cite{ren2022multi} further extended this line of work to multi-objective replanning by proposing Multi-Objective Path-Based D* Lite (MOPBD*). The method considers dynamic graphs with vector-valued edge costs, where objectives such as travel risk, time, or energy are optimized simultaneously. Instead of computing a single weighted path, MOPBD* maintains a Pareto-optimal set of paths and combines path-based expansion with incremental search. When newly detected obstacles or edge-cost changes modify the graph, the algorithm removes affected path labels and their descendants, regenerates inconsistent states, and incrementally updates the Pareto-optimal path set for replanning. An $\epsilon$-dominance variant is also introduced to approximate the Pareto front and improve efficiency. Li et al.~\cite{li2023bidirectional} proposed Bidirectional Lifelong Planning A* (BLPA*) that combines forward and backward searches to reuse previous search results when the environment changes. To improve replanning efficiency, the method constrains the forward search to the robot's perception range and performs a backward search for the remaining region.

\vspace{-3pt}
\subsection{Spatiotemporal Search-based Methods}

Assuming that dynamic obstacle trajectories are known or predicted, Phillips and Likhachev~\cite{phillips2011sipp} proposed Safe Interval Path Planning (SIPP), which defines the search state using a safe time interval within which the configuration remains collision-free. By using safe intervals instead of explicit timestep discretization, SIPP reduces the spatio-temporal search space while preserving completeness and time-optimality under its assumptions. Several variants have been proposed to improve efficiency and adaptability, including sub-optimal SIPP~\cite{phillips2011planning}, anytime SIPP~\cite{narayanan2012anytime}, generalized SIPP~\cite{gonzalez2012using}, and multi-objective SIPP~\cite{ren2022MO}. 

While SIPP represents safe intervals on a fixed graph, Cao et al.~\cite{cao2019dynamic} constructed a graph from the Delaunay triangulation of pedestrian positions. Each edge between two neighboring pedestrians defines a gate, a time-varying passage that the robot can cross. The method predicts pedestrian trajectories, computes each gate's width for safe passage, and applies the modified Timed A* algorithm to find a feasible channel formed by adjacent triangles connected through feasible gates. The selected channel is then converted into a continuous path using funnel-based optimization. Huang et al.~\cite{huang2025safe} constructed a dynamic connected visibility graph, where safe intervals are computed for graph edges to account for moving obstacles and low-order dynamic bounds. It then uses Uniform Temporal Visibility Deformation (UTVD) to identify spatial-temporally distinct initial paths, from which safe corridors are generated. Finally, B-spline trajectory optimization is performed within these corridors to obtain smooth, dynamically feasible trajectories to avoid static and moving obstacles.

\vspace{-3pt}
\subsection{Motion Primitive-based Methods}

These methods build short-horizon trajectories, called motion primitives, that satisfy motion constraints while avoiding collisions with obstacles, and then construct the optimal path using the evolution of these motion primitives.

Lin et al.~\cite{lin2021search} proposed a search-based online partial motion planner for car-like robot navigation in dynamic environments. The method generates motion primitives by discretizing time and control inputs, searches a state-time graph, and improves efficiency through primitive aggregation and pruning. It further accelerates dynamic collision checking by linearizing relative motion between the robot and moving obstacles and evaluating point-line distances. Chen et al.~\cite{chen2022rast} proposed Risk-Aware Spatio-Temporal (RAST) safety corridors for micro aerial vehicle navigation in dynamic environments. The method uses a particle-based dynamic map to predict future occupancy under uncertainties and defines collision risk over future time intervals. A risk-aware kinodynamic A* search~\cite{zhou2019robust} samples motion primitives and evaluates their risks to generate a reference path, from which spatiotemporal safety corridors are constructed for trajectory optimization. 

Beyond local primitive search, some methods integrate motion primitives into hierarchical frameworks with global guidance or trajectory refinement. Chen et al.~\cite{chen2023risk} proposed the Risk-Aware Sampling-based (RAS) algorithm, which combines local risk-aware trajectory generation with global reference path guidance. The method samples two-phase motion primitives to generate a local risk-aware trajectory, using a strict short-term risk constraint for immediate safety and a long-term risk cost to guide the robot toward safer regions. It allows the robot to follow a global reference path when it is safe, deviate to a local trajectory when the global path is blocked, and merge back to the global path once a safe connection becomes available. Qi et al.~\cite{qi2023hierarchical} proposed a hierarchical motion planning framework that generates smooth and feasible trajectories. The method first uses a partial-motion-planning-based search with spatiotemporal motion primitives, combined heuristics, and fast collision checking to obtain a coarse trajectory. It then applies multilayer optimization to smooth the path and speed profile, producing a dynamically feasible trajectory for avoiding static and moving obstacles. Wiman and Tiger~\cite{wiman2025safe} proposed Safe Lattice Planner (SLP), which uses a high-resolution lattice with the full primitive set for short-term collision-free planning and a low-resolution lattice with a reduced primitive set to provide longer-horizon guidance. If no complete collision-free path to the goal is available, the planner switches to a partial trajectory that remains safe while moving toward the goal.

\subsection{Summary of Search-based Methods}

Search-based methods provide structured and interpretable solutions by converting environmental changes into graph updates, temporal feasibility constraints, or motion-primitive evaluations. Incremental search methods are effective when environmental changes are local, since they can reuse previous search results and update only affected graph regions. However, their performance depends on the quality of the underlying spatial graph and may degrade when obstacle motion induces frequent or large-scale edge-cost changes. Spatiotemporal search methods improve temporal reasoning by incorporating safe intervals, time-varying gates, or spatiotemporal corridors, but their efficiency relies heavily on accurate obstacle prediction and compact temporal representations. Motion primitive-based methods further improve dynamic feasibility by embedding robot motion constraints into the search process, yet their effectiveness is tied to the design and resolution of the primitive set. Overall, search-based methods offer strong interpretability and reliable graph-level reasoning, but balancing discretization accuracy, computational efficiency, prediction uncertainty, and kinodynamic feasibility remains a major challenge in dynamic environments.

\section{Model Predictive Control Methods}
\label{mpc}

Model Predictive Control (MPC) is widely adopted for motion planning in dynamic environments because it integrates system dynamics, safety constraints, and time-varying objectives into a unified optimization framework. At each planning step, MPC solves a finite-horizon optimal control problem to predict future system behavior and update control inputs in real time. This receding-horizon strategy enables the robot to proactively adapt to environmental changes while maintaining feasibility and safety. To improve robustness and long-term task progress, recent MPC-based planners have incorporated risk-aware constraints, uncertainty-aware trajectory prediction, and hierarchical guidance mechanisms to address challenges such as dynamic-obstacle uncertainty and deadlock.

\subsection{Safety Constraint Formulation}
Several methods formulate safety constraints using the predicted positions and uncertainties of dynamic obstacles. Obstacles are commonly represented as convex polyhedra, such as cuboids, or as smooth differentiable surfaces, such as ellipsoids. Polyhedral obstacles are typically encoded as disjunctions of linear inequality constraints, which can be transformed into deterministic constraints~\cite{Blackmore2011}. However, this formulation can be computationally expensive in crowded  environments. In contrast, representing obstacles as differentiable surfaces allows collision avoidance to be expressed using smooth nonlinear constraints, avoiding the binary variables and numerous face constraints required by polyhedral formulations, thus reducing the number of constraints.  However, this strategy may produce conservative solutions.

\begin{figure}[t]
        \centering        \includegraphics[width=0.40\textwidth]{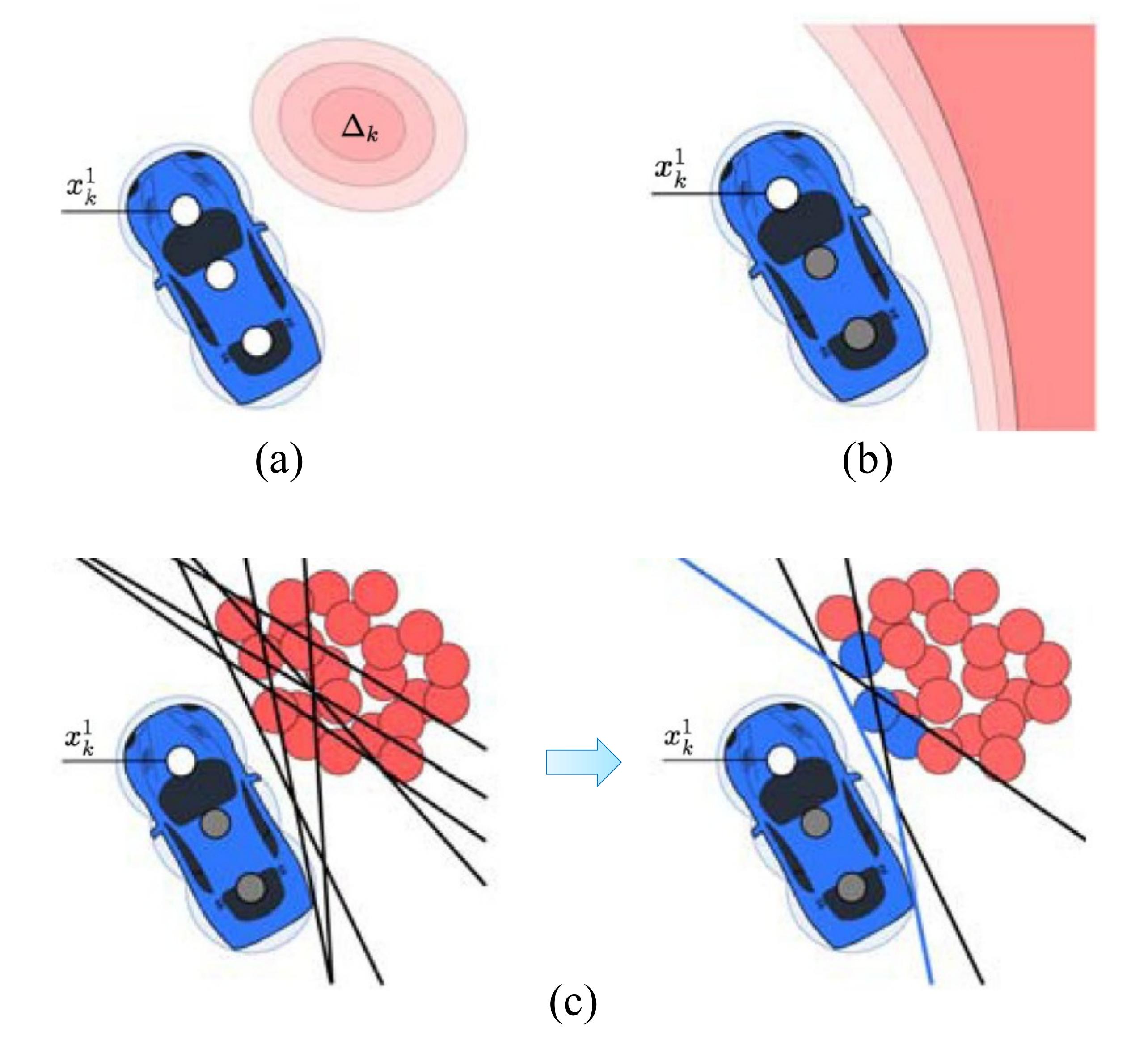}
    \caption{Different safety constraint formulations for dynamic obstacle with position uncertainty and ellipsoidal boundary~\cite{Groot2021}. (a) The red region illustrates the $1\sigma$ to $3\sigma$ interval of the collision uncertainty. (b) A linearized chance constraint is applied over the uncertainty region shown in (a). (c) The chance constraint is reformulated into a set of deterministic constraints by sampling (red circles), and a collision avoidance boundary (black line) is computed for each sample. To improve efficiency, only critical samples (blue circles) are retained.}\label{fig:safetyConstraint} 
    \vspace{-1em}
 \end{figure}

Brito et al.~\cite{Brito2019} developed the Local Model Predictive Contouring Control approach, which generates a collision-free trajectory to track a reference global path within a finite horizon. Dynamic obstacles are represented as ellipsoids, and the collision region is conservatively approximated in closed form using the Minkowski sum of an ellipsoid and the robot's circular footprint, as shown in Fig.~\ref{fig:safetyConstraint}a. Zhu and Alonso-Mora~\cite{Zhu2019} proposed the Chance-Constrained Nonlinear Model Predictive Control approach, which plans local trajectories while keeping the collision probability below a prescribed threshold. Dynamic obstacles are modeled as ellipsoids, and collision avoidance is formulated as chance constraints. These constraints are then linearized and converted into deterministic constraints for optimization, as illustrated in Fig.~\ref{fig:safetyConstraint}b.

Castillo et al.~\cite{castillo2020real} proposed a hybrid formulation for collision avoidance constraints. First, Polyhedral obstacles are used to obtain a closed-form approximation of disjunctive chance constraints. Then, a differentiable surface is constructed to provide a smooth and conservative bound around each polyhedral obstacle. To handle non-Gaussian uncertainties, Groot et al.~\cite{Groot2021} proposed the Scenario-based Model Predictive Contouring Control. It converts probabilistic collision constraints into deterministic constraints by sampling the uncertainty distribution to generate a set of scenarios. Each scenario defines a collision constraint at the sampled obstacle location, as illustrated in Fig.~\ref{fig:safetyConstraint}c. Mustafa et al.~\cite{mustafa2023probabilistic} extended this method by dynamically adjusting risk thresholds to balance safety and efficiency. Jian et al.~\cite{jian2023dynamic} used Control Barrier Functions to enforce the forward invariance of a predefined safe set, ensuring that the system state remains within the safety boundary by modifying control inputs in real time. 

Huang et al.~\cite{huang2025risk} introduced a Risk Euclidean Safety Metric that incorporates relative velocity into distance-based collision risk evaluation and formulates a corresponding Risk Control Barrier Function for planning. Saviolo et al.~\cite{saviolo2025reactive} computed a time-to-collision map and selected high-risk collision points, which are then used to define safety constraints in the MPC framework. Ryu and Mehr~\cite{ryu2024integrating} employed Conditional Value-at-Risk to approximate chance constraints in a convex and tractable manner. By focusing on the tail distribution of collision probabilities, this formulation provides tighter risk control than conventional chance constraints while remaining compatible with convex optimization frameworks.

\vspace{-3pt}
\subsection{Uncertainty-Aware Prediction}
Effective collision avoidance in dynamic environments depends not only on constraint formulation but also on the accuracy and reliability of dynamic-obstacle prediction. Many methods employ filter-based predictors, such as the Kalman filter~\cite{bar2004estimation}, to estimate future positions and associated uncertainties. These methods support uncertainty-aware planning but typically rely on assumptions about the underlying motion model and probability distribution, often Gaussian. Recently, neural network-based predictors~\cite{rasouli2019pie, rhinehart2019precog, alahi2016social, kothari2021human, salzmann2020trajectron} have gained popularity because they can model complex motion patterns without explicitly specifying dynamics or distributional forms. However, many neural predictors do not provide calibrated uncertainty estimates, which may lead to unsafe planning decisions when predictions are inaccurate.

\begin{figure}[t]
        \centering        \includegraphics[width=0.4\textwidth]{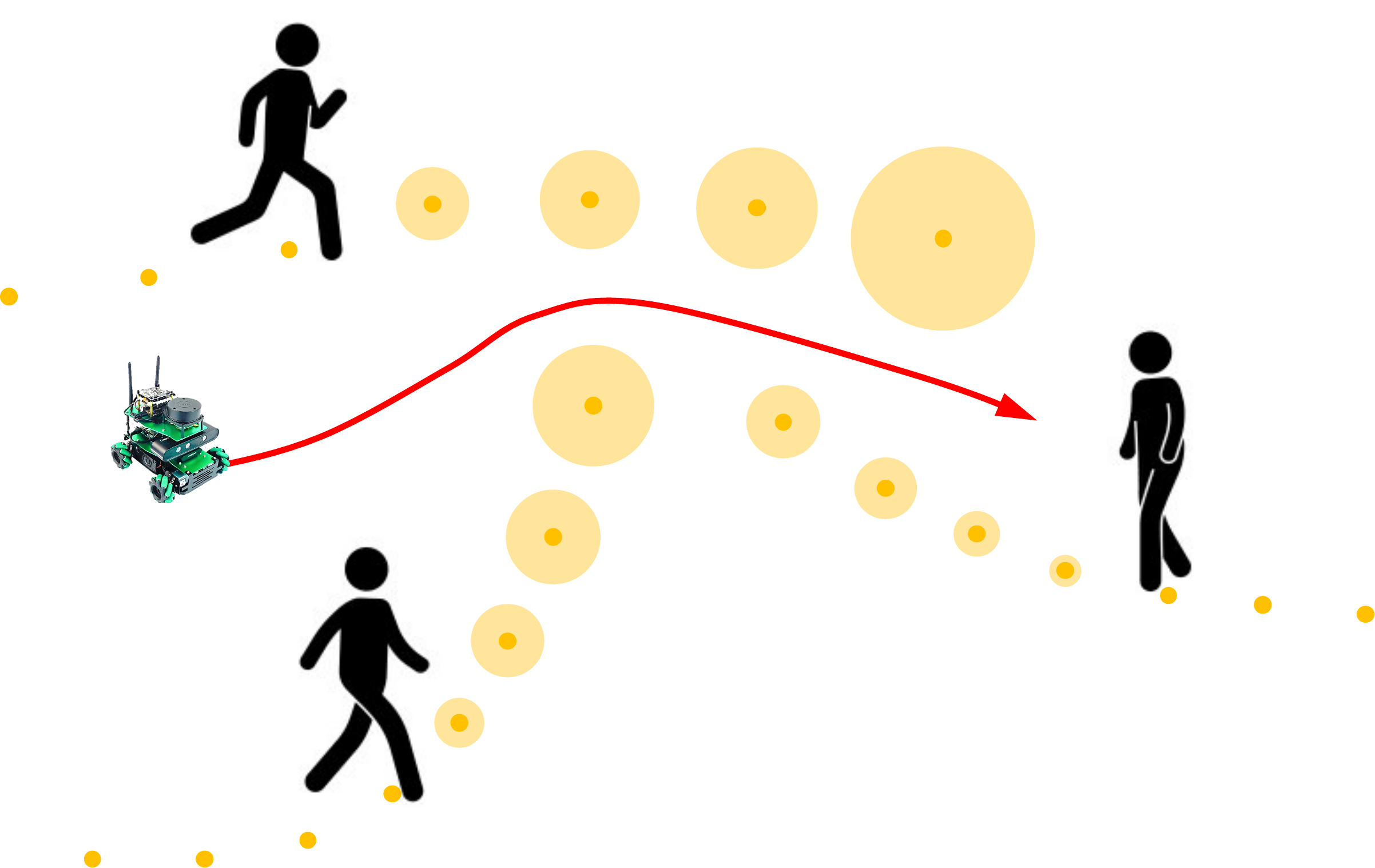}
    \caption{Conformal prediction~\cite{shafer2008tutorial} is used to generate valid prediction regions (orange circles) that quantify the uncertainty of neural network–based trajectory forecasts (orange dots)~\cite{Lindemann2023}.}\label{fig:conformalPred} 
    \vspace{-1em}
 \end{figure}

To address this limitation, Lindemann et al.~\cite{Lindemann2023} incorporate conformal prediction~\cite{shafer2008tutorial} to generate valid prediction regions that quantify the uncertainty of neural network-based trajectory forecasts, as shown in Fig.~\ref{fig:conformalPred}. These prediction regions are incorporated into an MPC framework, allowing the planner to reason about safety with formal uncertainty guarantees. This uncertainty quantification strategy is subsequently applied to a reinforcement learning-based planning method~\cite{Strawn2023}.

\subsection{Deadlock Mitigation}
\begin{figure}[t]
    \centering
    \subfloat[A neural network-based DRL planner generates subgoals for the MPC-based local planner~\cite{Brito2021}.]{
        \includegraphics[width=0.35\textwidth]{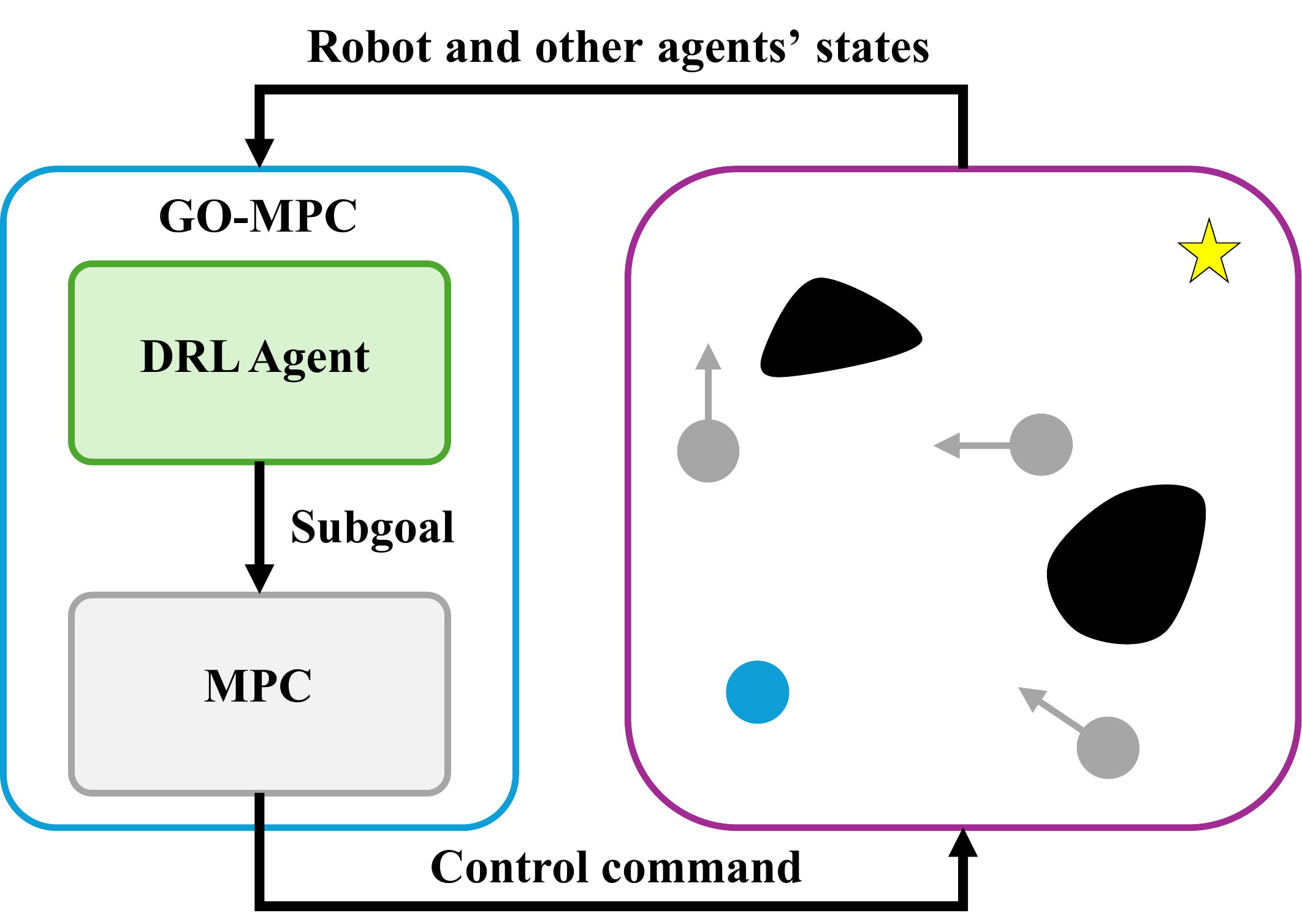}\label{fig:deadlockMitigate_part1}}\\
    \centering
    \subfloat[Multiple global paths are generated across distinct homotopy classes by a high-level planner to optimize and guide local trajectory generation~\cite{Groot2025}.]{
        \includegraphics[width=0.35\textwidth]{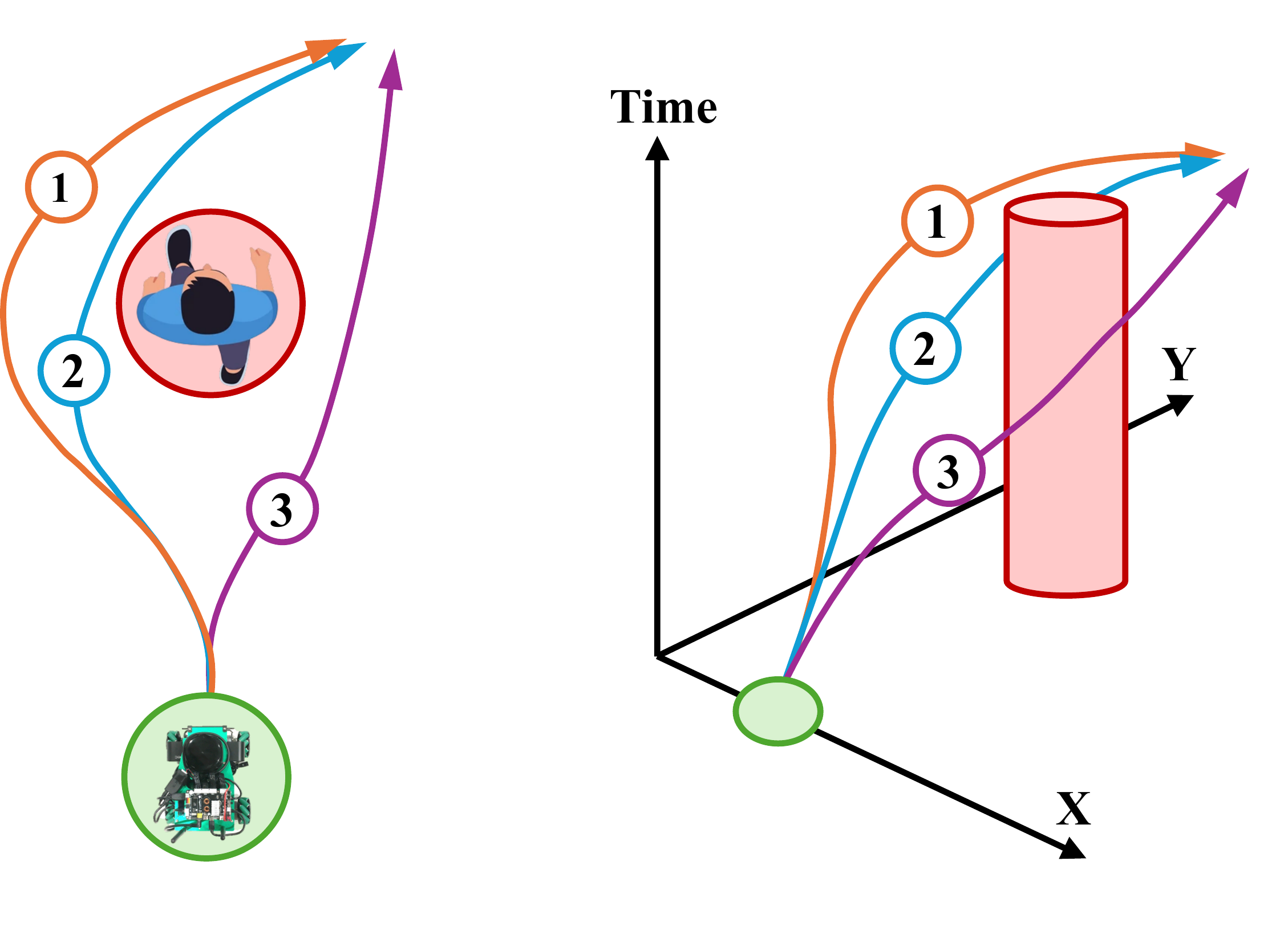}\label{fig:deadlockMitigate_part2}}\\
          \caption{Representative MPC-based methods with high-level guidance.}\label{fig:deadlockMitigate}
          \vspace{-1em}
\end{figure}

MPC-based planners may get trapped in deadlock situations due to local minima in the optimization landscape. To mitigate this issue, several methods introduce high-level guidance or modify the MPC cost function. Brito et al.~\cite{Brito2021} proposed Goal-Oriented MPC, which uses a learned policy to provide long-term guidance, as shown in Fig.~\ref{fig:deadlockMitigate}a. A deep neural network sets a subgoal for the MPC, and the MPC module generates control commands that satisfy different motion constraints while moving toward the suggested subgoal. 

Groot et al.~\cite{Groot2025} developed a Topology-driven MPC framework, which integrates a high-level planner based on visibility-probabilistic roadmap~\cite{simeon2000visibility} with a low-level MPC. In this method, multiple global paths are generated by the high-level planner across different homotopy classes by utilizing the topological structure of the free space, as shown in Fig.~\ref{fig:deadlockMitigate}b. Then, each path is exploited by a separate MPC-based local planner in parallel, and the trajectory with the lowest overall cost is selected for execution. Other approaches address deadlock by encouraging proactive behavior through cost-function design. Arul et al.~\cite{arul2023ds} designed a collision probability function that combines distance-to-obstacle and time-to-collision. Heuer et al.~\cite{heuer2023proactive} proposed a cost function based on multi-modal human motion predictions. Both methods encourage more assertive robot behavior while maintaining safety.

 \begin{figure*}[t]
    \centering        
    \includegraphics[width=0.90\textwidth]{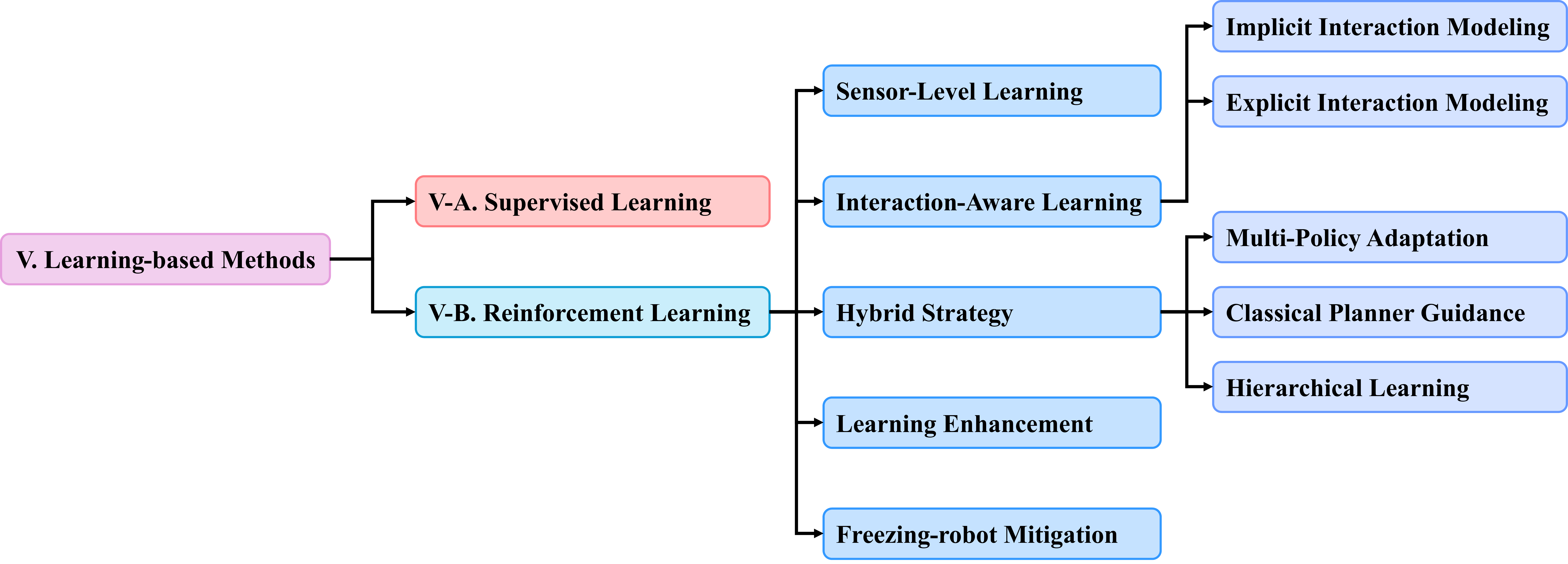}
    \caption{Organization and main topics of learning-based motion planning methods in dynamic environments.}\label{fig:RL_diagram} 
    \vspace{-1em}
 \end{figure*}

\subsection{Summary of Model Predictive Control Methods}

MPC-based methods provide a unified optimization framework for dynamic motion planning by considering system dynamics, safety constraints, and time-varying objectives within a receding horizon. Safety-constraint formulations improve collision avoidance by encoding dynamic obstacles as deterministic, chance-constrained, or risk-aware constraints. However, these formulations often introduce a trade-off between safety, conservatism, and computational tractability, especially in crowded environments with multiple uncertain obstacles. Uncertainty-aware prediction further improves robustness by incorporating probabilistic forecasts or calibrated prediction regions, but its effectiveness depends on the reliability of the prediction model and the quality of uncertainty estimation. High-level guidance and cost-function design help mitigate deadlock and improve long-horizon task progress, yet they introduce additional dependencies on global planners, learned policies, or carefully designed objectives. Overall, MPC-based methods provide strong capabilities for enforcing dynamic feasibility and safety constraints. However, their practical performance is often limited by the need to solve complex optimization problems in real time while accounting for uncertain obstacle predictions and long-horizon interactions.

\begin{table*}[t]{}
\footnotesize
\caption {Qualitative comparison of representative RL-based motion planning methods in dynamic environments.}\label{tab:rlbasedfeature}\vspace{-3pt}
\centering
\setlength\tabcolsep{6pt}
\begin{tabular}{l l l l l l} 
 \toprule
\specialrule{0.1em}{1pt}{1pt} 
\tabincell{l}{\textbf{Category}}
&\tabincell{l}{\textbf{Benefit}}
&\tabincell{l}{\textbf{Paper}}
&\tabincell{l}{\textbf{Year}}
&\tabincell{l}{\textbf{Key Technique}}
&\tabincell{l}{\textbf{Method Summary}}\\ 

\specialrule{0.1em}{1pt}{1pt} 
\multirow{5}{*}{\textbf{Sensor-Level Learning}} 
&\multirow{5}{*}{\begin{tabular}[l]{@{}l@{}}{Easy to implement and} \\ {enables fast inference}\end{tabular}} 
&\tabincell{l}{\cite{long2018towards}}
&\tabincell{l}{2018}
&\tabincell{l}{Parallel PPO}
&\begin{tabular}[l]{@{}c@{}}{Directly maps the raw LiDAR data} \\ {to collision-free steering commands}\end{tabular}\vspace{3pt}\\

&\multirow{5}{*}{}
&\tabincell{l}{\cite{huang2021towards}}
&\tabincell{l}{2021}
&\tabincell{l}{{Segmentation}\\{network}}
&\begin{tabular}[l]{@{}l@{}}{Presents a multi-modal sensor fusion} \\ {framework for the policy learning module}\end{tabular}\vspace{3pt}\\

&\multirow{5}{*}{}
&\tabincell{l}{\cite{Heuvel2024}}
&\tabincell{l}{2024}
&\tabincell{l}{{Attention}\\{module}}
&\begin{tabular}[l]{@{}l@{}}{Designs a spatiotemporal attention pipeline to} \\ {infer the scene dynamics from raw LiDAR data }\end{tabular}\\

\specialrule{0.05em}{1pt}{2pt}

\multirow{12}{*}{\tabincell{l}{\textbf{Interaction-Aware}\\\textbf{Learning}}} 
&\multirow{12}{*}{\begin{tabular}[l]{@{}l@{}}{Generates robust and} \\ {cooperative policies}\end{tabular}} 
&\tabincell{l}{\cite{chen2017decentralized}}
&\tabincell{l}{2017}
&\tabincell{l}{{Joint action}\\{learning}}
&\begin{tabular}[l]{@{}l@{}}{Introduces a joint state representation that} \\ {implicitly models human-robot interactions}\end{tabular}\vspace{3pt}\\

&\multirow{5}{*}{}
&\tabincell{l}{\cite{Chen2017social}}
&\tabincell{l}{2017}
&\tabincell{l}{Social norm}
&\begin{tabular}[l]{@{}l@{}}{Promotes social-aware behaviors} \\ {using a norm-inducing reward term}\end{tabular}\vspace{3pt}\\

&\multirow{5}{*}{}
&\tabincell{l}{\cite{everett2018motion}}
&\tabincell{l}{2018}
&\tabincell{l}{LSTM}
&\begin{tabular}[l]{@{}l@{}}{Employs LSTM to handle the observations} \\ {of an arbitrary number of
dynamic obstacles}\end{tabular}\vspace{3pt}\\

&\multirow{5}{*}{}
&\tabincell{l}{\cite{Chen2019}}
&\tabincell{l}{2019}
&\tabincell{l}{{Attention}\\{module}}
&\begin{tabular}[l]{@{}l@{}}{Models and aggregates both human-robot and} \\ {human-human interactions by attention module}\end{tabular}\vspace{3pt}\\

&\multirow{5}{*}{}
&\tabincell{l}{\cite{chen2020robot}}
&\tabincell{l}{2020}
&\tabincell{l}{GNN}
&\begin{tabular}[l]{@{}l@{}}{Models agents and their interactions as nodes} \\ {and edges using graph neural network (GNN)}\end{tabular}\vspace{3pt}\\

&\multirow{5}{*}{}
&\tabincell{l}{\cite{yang2023st}}
&\tabincell{l}{2023}
&\tabincell{l}{Transformer}
&\begin{tabular}[l]{@{}l@{}}{Captures interactions in each frame and} \\ {their evolution over time using transformer}\end{tabular}\\

\specialrule{0.05em}{1pt}{2pt}

\multirow{7}{*}{\textbf{Hybrid Strategy}} 
&\multirow{7}{*}{\begin{tabular}[l]{@{}l@{}}{Improves scalability}\\{and generalization}\end{tabular}} 
&\tabincell{l}{\cite{fan2020distributed}}
&\tabincell{l}{2020}
&\tabincell{l}{{Scenario-aware}\\{policy selection}}
&\begin{tabular}[l]{@{}l@{}}{Selects PID, RL, and safety-oriented policies} \\ {based on the complexity of surrounding scenario}\end{tabular}\vspace{3pt}\\

&\multirow{5}{*}{}
&\tabincell{l}{\cite{wang2020mobile}}
&\tabincell{l}{2020}
&\tabincell{l}{{Global-guided}\\{training scheme}}
&\begin{tabular}[l]{@{}l@{}}{Develops a hierarchical planning method that uses} \\ {a global path to guide the RL-based local planner}\end{tabular}\vspace{3pt}\\

&\multirow{5}{*}{}
&\tabincell{l}{\cite{Han2025}}
&\tabincell{l}{2025}
&\tabincell{l}{Residual DRL}
&\begin{tabular}[l]{@{}l@{}}{Learns corrective actions to adjust MPC planner}\end{tabular}\vspace{3pt}\\

&\multirow{5}{*}{}
&\tabincell{l}{\cite{jing2024two}}
&\tabincell{l}{2024}
&\tabincell{l}{Hierarchical RL}
&\begin{tabular}[l]{@{}l@{}}{Uses a high-level RL to determine subgoals} \\ {to guide low-level RL control policy}\end{tabular}\\

\specialrule{0.05em}{1pt}{2pt}

\multirow{8}{*}{\textbf{Learning Enhancement}} 
&\multirow{8}{*}{\begin{tabular}[l]{@{}l@{}}{Improves efficiency,}\\{robustness, and}\\{generalization}\end{tabular}} 
&\tabincell{l}{\cite{tai2018socially}}
&\tabincell{l}{2018}
&\tabincell{l}{Imitation learning}
&\begin{tabular}[l]{@{}l@{}}{Provides a data-efficient way to acquire}\\{policies by leveraging expert demonstrations}\end{tabular}\vspace{3pt}\\

&\multirow{5}{*}{}
&\tabincell{l}{\cite{perez2021robot}}
&\tabincell{l}{2021}
&\tabincell{l}{{Compositional}\\{learning framework}}
&\begin{tabular}[l]{@{}l@{}}{Trains policies in simple scenarios and}\\{transfers them  to more complex scenarios}\end{tabular}\vspace{3pt}\\

&\multirow{5}{*}{}
&\tabincell{l}{\cite{Liu2024}}
&\tabincell{l}{2024}
&\tabincell{l}{{Experience}\\{transfer}}
&\begin{tabular}[l]{@{}l@{}}{Accelerates learning process by}\\{transferring MPC experience data}\end{tabular}\vspace{3pt}\\

&\multirow{5}{*}{}
&\tabincell{l}{\cite{Sen2025}}
&\tabincell{l}{2025}
&\tabincell{l}{{Domain}\\{randomization}}
&\begin{tabular}[l]{@{}l@{}}{Generates diverse training scenarios}\end{tabular}\\

\specialrule{0.05em}{1pt}{2pt}

\multirow{8}{*}{\textbf{Freezing-robot Mitigation}} 
&\multirow{8}{*}{\begin{tabular}[l]{@{}l@{}}{Improves applicability}\\{in cluttered scenarios}\end{tabular}} 

&\tabincell{l}{\cite{sathyamoorthy2020frozone}}
&\tabincell{l}{2020}
&\tabincell{l}{{Freezing zone}\\{formulation}}
&\begin{tabular}[l]{@{}l@{}}{Computes a deviation to adjust the velocity}\\{output of RL to avoid entering freezing zone}\end{tabular}\vspace{3pt}\\

&\multirow{5}{*}{}
&\tabincell{l}{\cite{Zhu2023}}
&\tabincell{l}{2023}
&\tabincell{l}{{Modified reward}\\{function}}
&\begin{tabular}[l]{@{}l@{}}{Uses oriented capsules to enclose obstacles;}\\{adds a velocity-aware risk to reward function}\end{tabular}\vspace{3pt}\\

&\multirow{5}{*}{}
&\tabincell{l}{\cite{xie2023drl}}
&\tabincell{l}{2023}
&\tabincell{l}{{VO-enhanced}\\{reward function}}
&\begin{tabular}[l]{@{}l@{}}{Creates a reward term based on velocity obstacles}\\{(VO) to promote active collision avoidance}\end{tabular}\vspace{3pt}\\

&\multirow{5}{*}{}
&\tabincell{l}{\cite{Du2025}}
&\tabincell{l}{2025}
&\tabincell{l}{{Congestion-aware}\\{scheme}}
&\begin{tabular}[l]{@{}l@{}}{Leverages predicted congestion information to}\\{improve navigation policy in cluttered scenarios.}\end{tabular}\\

\bottomrule
\specialrule{0.1em}{1pt}{1pt}
\end{tabular}
\vspace{-1.0em}
\end{table*}

\section{Learning-based Methods}
\label{learnmethod}
Recent years have witnessed a growing interest in learning-based techniques for motion planning, particularly in dynamic and uncertain environments. In comparison to classical planners that often rely on a priori designed rules, explicit models, and carefully tuned parameters, learning-based methods can learn navigation behaviors from data or interactions. These methods have shown potential in adapting to human behaviors, handling uncertain observations, and generalizing across diverse scenarios. In this section, we review representative learning-based motion planning methods, focusing on supervised and reinforcement learning. Fig.~\ref{fig:RL_diagram} shows the organization and main topics reviewed in this section.

\vspace{-3pt}
\subsection{Supervised Learning}
Supervised learning-based methods formulate motion planning as a regression or classification task, mapping sensor data to expert-labeled control actions. Long et al.~\cite{long2017deep} employed deep neural networks trained via supervised learning to map LiDAR data directly to collision-free control commands. Pokle et al.~\cite{pokle2019deep} used multiple
convolutional neural networks (CNNs) to learn multi-modal high-level features from raw sensor data. A CNN-based local planner fuses these features to generate control commands. Xie et al.~\cite{xie2021towards} developed a CNN-based method that fuses a short history of LiDAR data with the position and velocity information of nearby agents to produce control commands. Qin et al.~\cite{Qin2021} introduced a deep imitation learning framework that takes raw sensor data and agent states as input, and generates socially compliant navigation behaviors. Hong et al.~\cite{hong2023obstacle} introduced a conditional variational autoencoder framework that learns temporary target positions from demonstrations to enable effective collision avoidance with fast-moving pedestrians. 

In summary, supervised learning-based methods learn navigation policies from expert demonstrations. However, they rely heavily on large and diverse training datasets, and their limited generalization to unseen scenarios restricts their effectiveness in complex real-world environments.

\vspace{-3pt}
\subsection{Reinforcement Learning} 
\label{sec:RL}
Reinforcement Learning (RL) has emerged as a powerful paradigm for navigation in dynamic environments. By learning policies through trial-and-error while interacting with the environment, RL enables robots to make sequential decisions that optimize long-term performance. This section provides a structured overview of recent RL-based navigation methods categorized by input modalities, modeling approaches for agent interactions, integration techniques with classical planning, learning efficiency enhancements, and freezing robot problem solutions. Table~\ref{tab:rlbasedfeature} summarizes the key features of representative RL-based methods.

\vspace{3pt}
\subsubsection{Sensor-Level Learning}
These methods directly learn navigation policies from raw sensor inputs without relying on explicit environmental modeling or agent-level reasoning. By leveraging end-to-end learning architectures, these methods bypass the need for traditional pipelines involving perception, mapping, and tracking. This simplicity makes them attractive for real-time navigation tasks, especially in dynamic environments where quick reactions are crucial.

Long et al.\cite{long2018towards} trained navigation policies for a multi-robot system using the Proximal Policy Optimization (PPO) algorithm\cite{schulman2017proximal}. Fan et al.~\cite{Fan2019} extended this work by addressing the navigation lost problem caused by localization uncertainty. Their framework includes two modes: a normal mode that uses an RL-based planner and a recovery mode that guides the robot to landmark-rich areas for re-localization. Huang et al.~\cite{huang2021towards} presented a multi-modal sensor fusion framework that combines RGB images and LiDAR data to enhance environmental perception. Their method trains a segmentation network using supervised learning to convert raw RGB images into binary segmentation maps indicating traversable and non-traversable areas. The policy module then takes the segmentation map along with LiDAR data as input for training. By integrating complementary visual and geometric information, this approach provides a more comprehensive understanding of the environment compared to single-modality methods, thereby improving the robustness of the learned navigation policy in complex scenes. 

Han et al.~\cite{han2022deep} proposed a sensor fusion method that integrates 2D LiDAR data and RGB images to estimate a dense depth map. This map is then compressed into a 2D minimal depth representation and fused with the original LiDAR data to provide enhanced depth perception for policy learning. Heuvel et al.~\cite{Heuvel2024} designed a spatiotemporal attention pipeline to infer the scene dynamics from raw LiDAR data without explicitly tracking the dynamic obstacles. Their method employs a spatial attention module to identify relevant LiDAR sectors and a temporal attention module to aggregate information across time. The attention-weighted representation is fed into a network that is trained via the deep deterministic policy gradient (DDPG) algorithm~\cite{lillicrap2015continuous}. 

In summary, sensor-level methods offer a lightweight and efficient solution for robot navigation, particularly in scenarios where fast decision-making is essential. However, their reliance on raw sensor inputs also makes them sensitive to noise. Moreover, the absence of explicit modeling of agent behaviors limits their ability to navigate effectively in socially complex or densely populated environments, where understanding and reasoning about interactions are essential.

\vspace{3pt}
\subsubsection{Interaction-Aware Learning}
\label{InteractionAwareness}
In crowded environments, safe and reliable navigation requires an understanding of the surrounding agents’ behaviors and intentions. This challenge is addressed by Interaction-aware methods, which explicitly model agent-agent interactions and incorporate this relational model into the learning process. By leveraging structured representations of interactions, these methods enable socially compliant, cooperative, and anticipatory navigation behaviors that go beyond the capabilities of sensor-level approaches.

\vspace{3pt}
a) \textit{Implicit Interaction Modeling}: Chen et al.~\cite{chen2017decentralized} used a joint state representation to capture the states of the robot and the humans, thereby implicitly modeling human-robot interactions. Building upon this work, Chen et al.~\cite{Chen2017social} promoted socially aware behaviors (e.g., passing from the right) by incorporating an induced norm penalty into the reward function. Later in~\cite{everett2018motion}, the authors employed long short-term memory (LSTM)~\cite{hochreiter1997long} to encode the variable-sized joint state of surrounding agents into a fixed-length representation, enabling the robot to handle an arbitrary and dynamically changing number of humans over time. Instead of manually specifying social rules, Kathuria et al.~\cite{kathuria2025learning} proposed an inverse reinforcement learning framework to learn social navigation reward maps from few-shot expert demonstrations. The method uses scene geometry, human-robot trajectory histories, human velocity and heading, and the robot goal as inputs. The learned map is then used to generate reference trajectories executed by a local controller to enable implicit yielding and deadlock avoidance.

\begin{figure}[t]
        \centering        \includegraphics[width=0.4\textwidth]{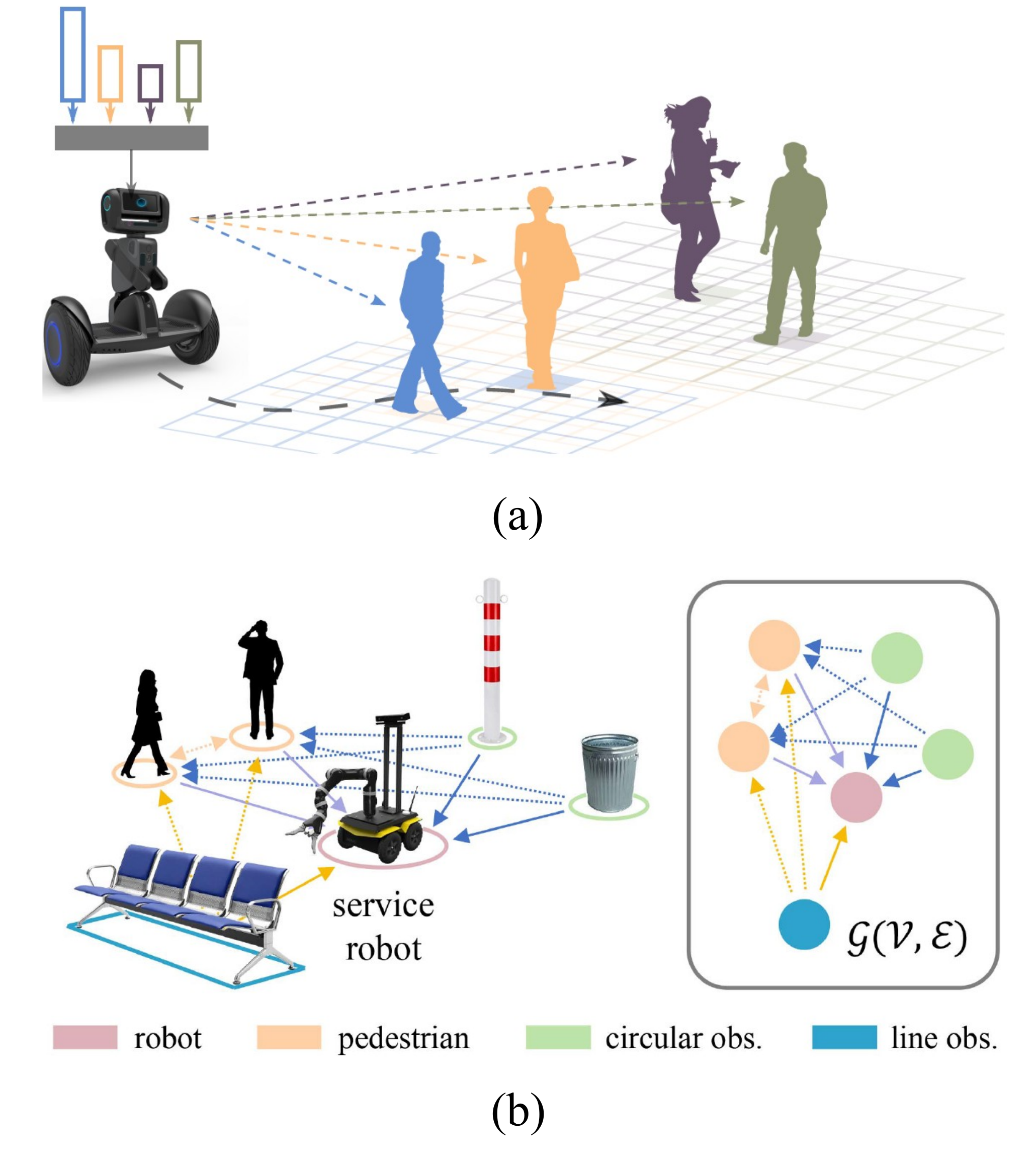}
    \caption{Representative agent-agent interaction modeling methods: (a) attention model~\cite{Chen2019} and (b) graph neural network~\cite{Liu2024}.}\label{fig:interactionModel} 
    \vspace{-0.5em}
 \end{figure}

\vspace{3pt}
b) \textit{Explicit Interaction Modeling}: The above methods focus solely  on human-robot interactions, overlooking the effects of interactions between humans. To address this limitation, Chen et al.~\cite{Chen2019} explicitly modeled the human-robot interactions
and encoded the human-human interactions
using coarse-grained local maps. Fig.~\ref{fig:interactionModel}a shows the interactions between the robot and each human, which are aggregated in the attention pooling module. The resulting aggregated representation informs the value network used for policy learning. This structure enables the robot to reason about both direct and indirect interactions in dense crowds. However, this method does not explicitly model static obstacles, instead treating them as stationary humans, which can lead to sub-optimal decisions. To overcome this limitation, Liu et al.\cite{LiuLucia2020} enhanced the framework by processing static and dynamic obstacles separately, allowing for a more natural reaction. 

To capture structured relational dependencies, recent methods~\cite{chen2020relational,chen2020robot,liu2021decentralized} encoded agents and their interactions as nodes and edges of a Graph neural network (GNN)~\cite{kipf2016semi}. These GNN-based methods were extended later for performance enhancement. For instance, Jiang et al.\cite{jiang2024learning} modeled asymmetric interactions between the robot and humans. Liu et al.~\cite{Liu2024} modeled heterogeneous interactions between the robot and different types of obstacles, as seen in Fig.~\ref{fig:interactionModel}b. Zhou et al.~\cite{Zhou2025} proposed a similar method to capture heterogeneous interactions. Lu et al.~\cite{Lu2025} modeled spatio-temporal interactions capturing  relational features. Recently, Transformer-based architectures~\cite{vaswani2017attention} have also gained attention. Yang et al.~\cite{yang2023st} proposed a transformer-based method that captures agent-to-agent interactions in each frame and their evolution over time. The resulting representation is used in a value-based RL framework.

In summary, by explicitly reasoning about the behaviors and intentions of surrounding agents, interaction-aware methods produce navigation policies that are more robust, socially aware, and cooperative than those based solely on raw sensor data. However, these benefits often come with trade-offs, including increased model complexity, higher computational demands, and a reliance on accurate perception and interaction modeling. Moreover, their performance can degrade in the presence of noisy or incomplete agent state information. Nevertheless, interaction-aware approaches are indispensable for safe, effective, and socially acceptable navigation in densely populated and dynamic environments.

\begin{figure}[t]
        \centering        \includegraphics[width=0.40\textwidth]{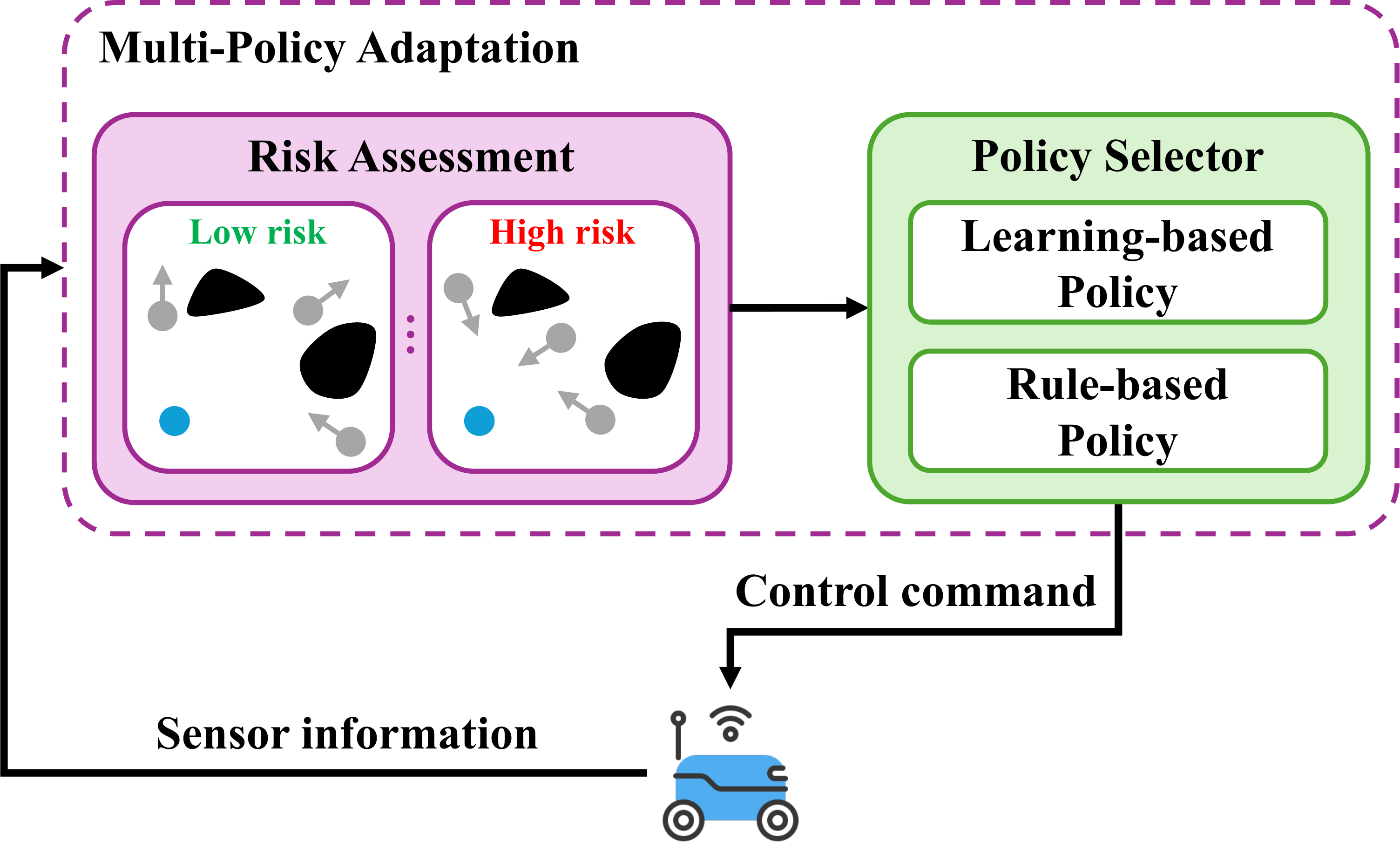}
    \caption{Multi-policy adaptation strategy: sensor information is used to assess the risk level, based on which the robot selects among learning-based and rule-based policies to generate control commands.}\label{fig:multipolicy} 
    \vspace{-1em}
 \end{figure}

\vspace{6pt}
\subsubsection{Hybrid Strategy} 
RL-based local planners often struggle with sparse rewards in large environments and limited generalization due to the sim-to-real gap. To address these limitations, hybrid strategies have emerged as a promising solution.

\vspace{3pt}
a) \textit{Multi-Policy Adaptation}: As shown in Fig.~\ref{fig:multipolicy}, a multi-policy architecture switches between a learning-based policy and a rule-based policy based on scenario complexity or risk level. Fan et al.~\cite{fan2020distributed} introduced a scenario-aware policy adaptation method that explicitly classifies navigation contexts into simple, complex, and emergent categories based on the robot's proximity to dynamic obstacles and a user-defined risk threshold. Depending on the scenario, the robot selects from three navigation strategies: a PID controller for simple cases, an RL-based planner for complex situations, and a safety-oriented policy for emergencies. Similar approaches are also adopted in~\cite{semnani2020multi,wu2023risk,matsumoto2024crowd,Qin2024}. 

\begin{figure}[t]
        \centering        \includegraphics[width=0.40\textwidth]{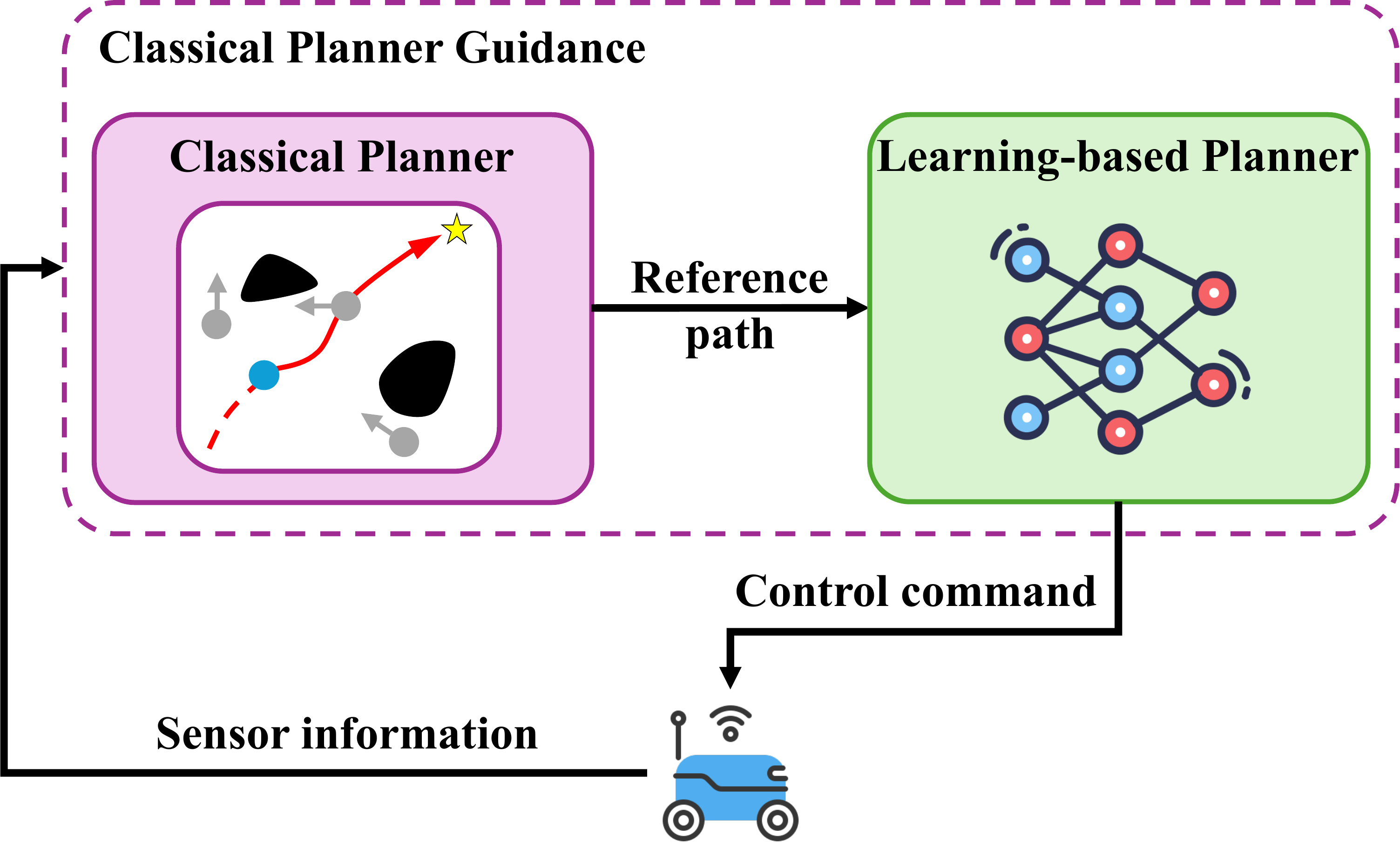}
    \caption{Classical planner strategy: a classical planner provides a reference path, while a learning-based local planner uses this path and sensor information to generate control commands.}\label{fig:classicalguidance} 
    \vspace{-1em}
 \end{figure}

\vspace{3pt}
b) \textit{Classical Planner Guidance}: As shown in Fig.~\ref{fig:classicalguidance}, Wang et al.\cite{wang2020mobile} developed a hierarchical path planning algorithm that leverages global guidance to train an RL-based local planner. Specifically, the A* algorithm\cite{hart1968formal} is used to generate a globally optimal path, which serves as a high-level reference. The RL-based local planner then learns to generate collision-free actions by considering both the global path and local environmental observations, including static and dynamic obstacles. Similar global-guided training schemes are also explored in~\cite{guldenring2020learning,angulo2023policy,Zhu2025}. Han et al.~\cite{Han2025} used Residual Deep RL ~\cite{johannink2019residual} to learn corrective actions on top of the MPC, enhancing control robustness and safety.

\begin{figure}[t]
        \centering        \includegraphics[width=0.40\textwidth]{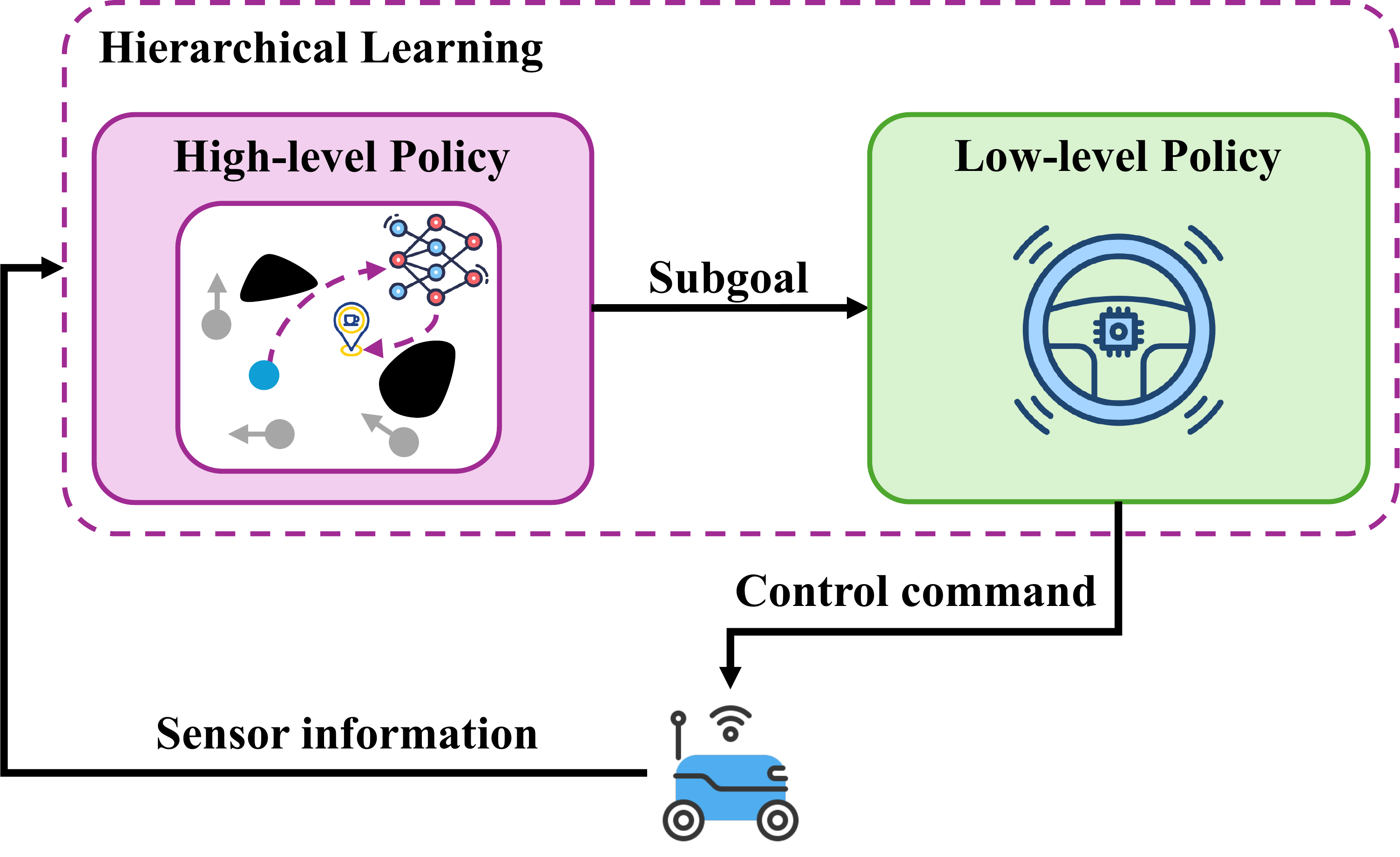}
    \caption{Hierarchical learning strategy: a high-level policy generates subgoals from sensor information, and a low-level policy produces control commands to navigate toward the selected subgoals.}\label{fig:hierarchicalRL} 
    \vspace{-1em}
 \end{figure}

\vspace{3pt}
c) \textit{Hierarchical Learning}: As shown in Fig.~\ref{fig:hierarchicalRL}, Jing et al.~\cite{jing2024two} proposed a hierarchical RL framework, where the high-level RL policy generates subgoals that guide the robot to avoid dense crowds while progressing towards the final destination, and the low-level network produces control actions to navigate towards these subgoals. Chen et al.~\cite{chen2024environmental} adopted a similar hierarchical framework, where a high-level network assesses environmental complexity to determine appropriate navigation behaviors (e.g., obstacle avoidance or goal pursuit), while a low-level network generates the corresponding control commands. Similar hierarchical learning methods are also explored in~\cite{ZhuWei2023, Gao2024, qin2024non,gao2025hierarchical}.

In summary, Hybrid strategies offer a balanced framework that improves the scalability and generalization of navigation policies. By switching between policies based on scenario complexity or combining different planners with reinforcement learning, these methods enable more effective decision-making across diverse environments while addressing safety and performance concerns. Although they introduce additional architectural complexity, hybrid strategies represent a practical and scalable approach for deploying learning-based navigation systems in real-world, safety-critical applications. 

\vspace{3pt}
\subsubsection{Learning Enhancement} 
RL-based navigation policies often require extensive interactions with the environment and suffer from poor generalization when deployed in unseen dynamic scenarios. To address these limitations, learning enhancement techniques introduce additional supervision, structured training processes, or more diverse interaction data to improve training efficiency, robustness, and generalization.

One line of work leverages demonstrations or expert knowledge to guide policy learning. Tai et al.~\cite{tai2018socially} used the generative adversarial imitation learning strategy~\cite{ho2016generative} to learn socially compliant navigation policies. Xu and Karamouzas~\cite{xu2021human} leveraged knowledge distillation~\cite{hinton2015distilling} to shape the reward function based on suboptimal expert policies extracted via behavior cloning from human trajectory demonstrations. Liu et al.~\cite{Liu2024} further used an offline MPC-based planner to generate demonstration data, which bootstraps RL training and reduces the number of required interaction steps.

Another line of work improves scalability through compositional learning. Perez et al.~\cite{perez2021robot} first trained policies in simple layouts, such as corridors, and then transferred them to more complex environments, enabling better adaptability across different navigation scenarios. In parallel, diversity-aware training methods expose the policy to richer dynamic interactions. Sen et al.~\cite{Sen2025} used domain randomization with optimal reciprocal collision avoidance~\cite{ocra} to generate diverse pedestrian behaviors. Wu et al.~\cite{Wu2025} learned a diversity-aware crowd model that produces varied yet realistic human navigation patterns. By increasing the diversity of training samples, these methods improve the robustness and generalization of RL policies in complex environments.

In summary, Learning enhancement methods improve RL-based motion planning mainly by strengthening the training process. Demonstration-guided learning accelerates policy acquisition; curriculum and compositional learning improve scalability across environments; and diversity-aware training enhances robustness to unseen human behaviors and dynamic interactions. These techniques make RL policies practical for real-world navigation; however, their performance depends on the quality of demonstrations, the diversity of training, and the similarity between simulated and real-world environments.

 \begin{table*}[t]
\centering
\caption{Qualitative comparison of representative classical local planning methods in dynamic environments.}\vspace{-3pt}
\label{tab:classical_local_planners}
\centering
\setlength\tabcolsep{5pt}
\begin{tabular}{l l l l}
\toprule
\specialrule{0.1em}{1pt}{1pt} 
\tabincell{l}{\textbf{Method Family}} 
&\tabincell{l}{\textbf{Main Principle}} 
&\tabincell{l}{\textbf{Strengths}} 
&\tabincell{l}{\textbf{Limitations}} \\
\toprule

\tabincell{l}{\textbf{Velocity Obstacle}} 
&\tabincell{l}{{Constructs velocity-space constraints that identify}\\{robot velocities leading to future collisions, and}\\{selects safe velocities outside these regions}} 
&\tabincell{l}{{Explicitly captures relative motion}\\{and is suitable for multi-agent}\\{collision avoidance}} 
&\tabincell{l}{{Requires accurate agent-state}\\{estimation}} \\

\specialrule{0em}{2pt}{2pt}
\tabincell{l}{\textbf{Potential Field}} 
&\tabincell{l}{{Generates attractive forces toward the goal and}\\{repulsive forces around obstacles, guiding the}\\{robot by the resulting force field}} 
&\tabincell{l}{{Easy to implement and capable}\\{of producing smooth motions}} 
&\tabincell{l}{{Prone to local minima}} \\

\specialrule{0em}{2pt}{2pt}
\tabincell{l}{\textbf{Dynamic Window}}
&\tabincell{l}{{Samples admissible velocity commands under}\\{robot dynamic constraints and selects the best}\\{command according to an objective function}} 
&\tabincell{l}{{Directly accounts for robot dynamic}\\{constraints and produces feasible}\\{velocity commands}}
&\tabincell{l}{{Limited to handcrafted cost}\\{functions}} \\
\bottomrule
\specialrule{0.1em}{1pt}{1pt} 
\end{tabular}
\vspace{-1.0em}
\end{table*}

\vspace{3pt}
\subsubsection{Freezing-robot Mitigation} 
In cluttered or crowded environments, RL-based planners often suffer from the freezing robot problem, where the robot comes to a stop or moves overly conservatively because most nearby actions are considered risky (e.g., surrounded by humans and no escape route is available). Existing methods mitigate this issue by encouraging more proactive behaviors through freezing-zone avoidance, risk-aware reward shaping, velocity-obstacle constraints, or congestion-aware coordination.

One line of work explicitly identifies regions or conditions that may cause freezing. Sathyamoorthy et al.~\cite{sathyamoorthy2020frozone} predicted pedestrian trajectories and constructed a Potential Freezing Zone (PFZ), which indicates regions where the robot is likely to freeze or obstruct human motion. The RL policy of the robot then adjusts the velocity output to avoid entering such regions. Du et al.~\cite{Du2025} addressed freezing from a congestion-aware perspective. Robots share their estimated arrival times at hallways, select hallway goals based on expected waiting time, and use a congestion predictor to determine when to enter a hallway, thereby avoiding heavily congested trajectories.

Another group of methods reduces excessive conservatism through risk-aware representation and reward design. Zhu et al.~\cite{Zhu2023} represented surrounding agents using oriented bounding capsules, which better capture their geometric and kinematic properties. The reward function incorporates minimum separating distance, state-dependent orientation, and velocity-related collision risk, enabling the policy to distinguish between slow- and fast-moving agents and respond more adaptively. Yang et al.~\cite{yang2023rmrl} relaxed hard collision penalties during training and introduced a probabilistic risk-aware reward based on predicted trajectories of dynamic agents. By treating collision risk as a soft constraint, the policy is encouraged to generate more proactive and flexible behaviors in dense crowds. Han et al.~\cite{han2022reinforcement} incorporated velocity obstacles~\cite{fiorini1998motion} into the reward function to penalize unsafe velocity choices. Xie and Dames~\cite{xie2023drl} formulated a velocity-obstacle-based reward term to promote active collision avoidance, together with a goal-attraction component that preserves task progress.

In summary, the freezing-robot mitigation methods prevent RL policies from becoming overly conservative in dense, dynamic environments. Freezing-zone and congestion-aware methods identify risky regions before the robot enters them, while risk-aware and velocity-obstacle-based reward designs encourage safer yet more proactive actions. These methods improve navigation in cluttered scenes, but their effectiveness depends on reliable prediction, appropriate risk modeling, and the ability to balance safety with forward progress.

\subsection{Summary of Learning-based Methods}

Learning-based methods provide data-driven solutions for motion planning in dynamic environments by learning navigation policies, interaction models, or auxiliary planning modules from demonstrations, experience, or simulation. Supervised learning methods offer a direct mapping from sensor observations or agent states to control commands, but their performance is strongly limited by the coverage and quality of expert-labeled data. RL-based methods improve autonomy by learning through interaction, yet different representations lead to different trade-offs. Sensor-level policies enable fast inference from raw observations but lack explicit reasoning about agent behaviors. Interaction-aware methods improve social compliance and cooperation by modeling human-robot and human-human interactions, but they depend on accurate perception and introduce higher model complexity. Hybrid strategies combine learned policies with classical planners, controllers, or high-level guidance to improve scalability and safety, although they also increase architectural complexity and require careful coordination between modules. Learning enhancement and freezing-robot mitigation methods further improve training efficiency and robustness, but their effectiveness still depends on simulation fidelity, scenario diversity, reward design, and risk modeling. Overall, learning-based methods offer strong adaptability in complex dynamic environments, but improving generalization, interpretability, safety verification, and sim-to-real transfer remains a central challenge for real-world deployment.

\section{Other Classical Methods}
\label{other}
This section reviews three representative classical methods that have been widely applied to motion planning in dynamic environments based on velocity obstacles, potential fields, and dynamic windows.  Although these methods were introduced decades ago, they remain highly relevant in practice. 
Their computational efficiency and ease of implementation make them attractive for resource-constrained robots such as service robots, warehouse robots, and aerial robots.  Moreover, they are often integrated as local planners or fallback modules in modern hybrid frameworks, complementing the learning-based or sampling-based  strategies.  Thus, these classical approaches continue to play a valuable role in practical deployments, despite the rise of more advanced planning paradigms. Table~\ref{tab:classical_local_planners} summarizes their main principle, strengths, and limitations.

\vspace{-6pt}
\subsection{Velocity Obstacle-based Methods}
Fiorini and Shiller~\cite{fiorini1998motion} introduced the concept of Velocity Obstacles (VO), which defines the set of relative velocities that would lead to collisions between a robot and surrounding agents. At each timestep, the robot selects a velocity outside of the VO set to ensure collision-free motion. To address the oscillatory behaviors that may arise when two agents react simultaneously, Van Den Berg et al.\cite{van2008reciprocal} proposed Reciprocal Velocity Obstacles (RVO), which assumes equal responsibility for collision avoidance between agents. This is further refined by Snape et al.\cite{snape2011hybrid}, who introduced Hybrid RVO (HRVO) to mitigate such oscillations. Liu et al.\cite{LiuZhihao2024} relaxed the reciprocity assumption by using Deep RL to simultaneously infer the responsibility for avoidance and determine escape velocities for improving navigation in crowded scenes. Martinez et al.\cite{Martinez2025} generalized this method by assuming the degree of cooperation from other agents is unknown. They proposed an adaptive opinion dynamics framework that enables robots to infer cooperation levels from onboard observations and thereby adjust their behaviors in real time. Subsequent studies extended the VO framework to accommodate various robot dynamics~\cite{van2011reciprocal,bareiss2013reciprocal,snape2010smooth,zhao2022solving}. 

VO-based methods perform well in structured or moderately dynamic environments where agents' states can be accurately perceived and their behaviors are predictable. They are computationally efficient for real-time implementation; however, their performance may degrade in crowded or unstructured environments, requiring integration with learning-based or traditional strategies to enhance robustness and flexibility.

\subsection{Potential Field-based Methods}
Traditional potential field-based methods construct an attractive potential towards the goal and repulsive potentials around the obstacles. The robot motion is guided by the resultant force, which is typically derived from the negative gradient of the potential field. To address dynamic environments, Ge and Cui~\cite{ge2002dynamic} defined attractive and repulsive potentials using the relative position and velocity between the robot, target, and obstacles. Chiang et al.~\cite{chiang2015path} combined sampling-based planning with potential fields, where a collision-free global path provides intermediate goals and the potential field planner handles local avoidance of dynamic obstacles. Malone et al.~\cite{malone2017hybrid} used stochastic reachable sets to construct repulsive fields for obstacles with uncertain motion. Boldrer et al.~\cite{boldrer2020socially} combined attractive potential fields with limit cycles around dynamic obstacles to generate smooth avoidance behaviors. Huber et al.~\cite{huber2022avoiding} constructed a converging vector field within a bounded volume to guide the robot around obstacles while ensuring convergence to the goal. Wang et al.~\cite{wang2026enhanced} introduced an attractive force with terminal acceleration to avoid terminal local minima, a 3D vortex-based repulsive force to guide safer obstacle circumvention, and an emergency-deflection mechanism to escape intermediate local minima.

Overall, potential field-based methods are simple and effective for smooth reactive navigation, but they remain susceptible to local minima in cluttered or crowded environments.

\begin{figure*}[t]
        \centering        \includegraphics[width=0.84\textwidth]{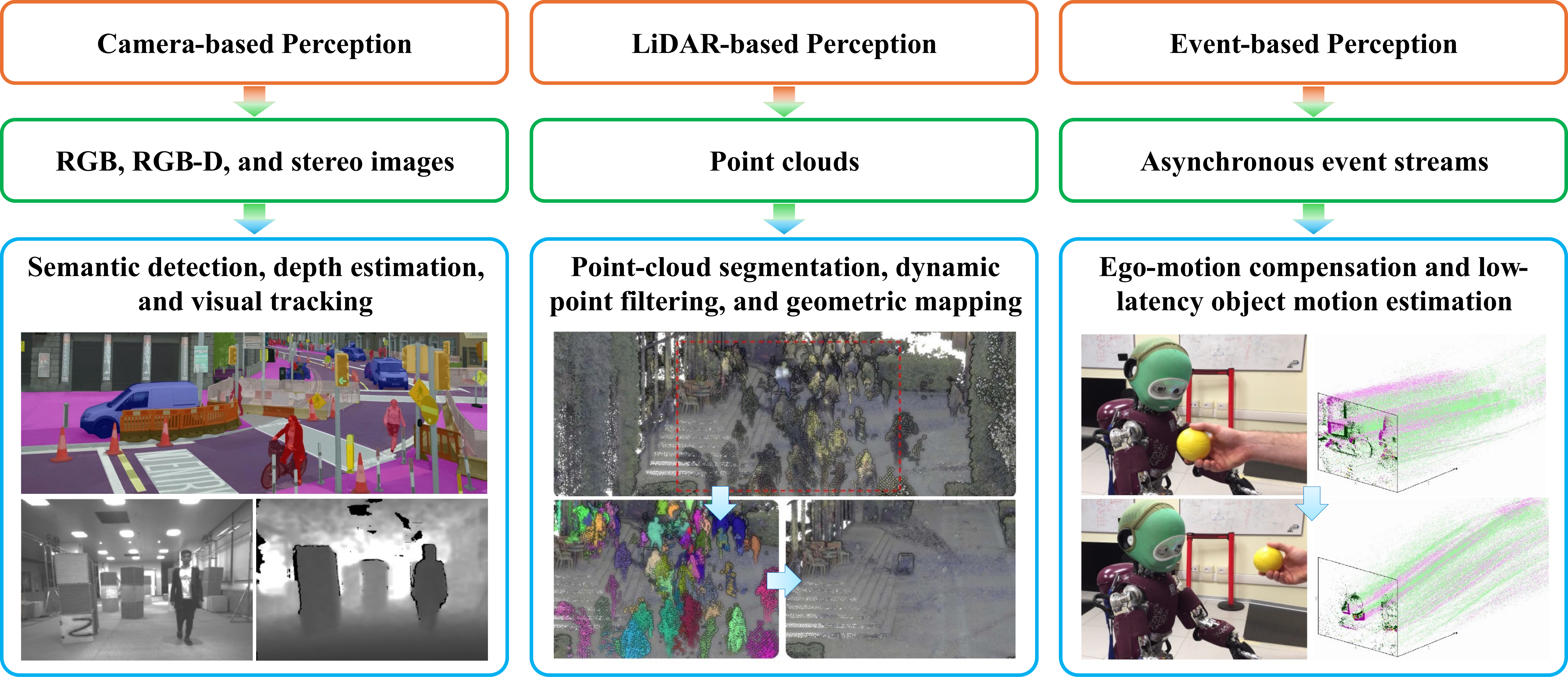}
    \caption{Overview of dynamic perception modalities, including camera-based~\cite{wang2021autonomous}, LiDAR-based~\cite{duberg2024dufomap}, and event-based perception~\cite{glover2017robust}.}\label{fig:perception} 
 \vspace{-0.5em}
 \end{figure*}

\vspace{-6pt}
\subsection{Dynamic Window-based Methods}
Dynamic Window Approaches (DWAs)~\cite{fox1997dynamic} optimize velocity commands by evaluating candidate actions given the robot's dynamic constraints. To handle dynamic environments, Missura and Bannewitz\cite{missura2019predictive} extended DWA by modeling moving obstacles as dynamic polygons. Chang et al.\cite{chang2021reinforcement} enhanced DWA by using tabular Q-learning to automatically tune the cost function weights. Patel et al.\cite{patel2021dwa} replaced the handcrafted cost function with a neural network that directly selects the best control action. Jian et al.\cite{jian2023long} proposed a long-term version of DWA, where an initial path is generated and refined using the Elastic Band method~\cite{quinlan1993elastic}. Yasuda et al.\cite{yasuda2023safe} adapted DWA to systems with uncertainty, ensuring safer navigation. Dobrevski and Sko{\v{c}}aj\cite{dobrevski2024dynamic} used a neural network to predict the DWA cost function parameters based on recent observations, enabling better handling of moving obstacles. Zhang et al.~\cite{zhang2025gradient} augmented DWA with a gradient-aware collision cost derived from obstacle-distance fields to anticipate potential collisions.

\vspace{-6pt}
\section{Dynamic Perception}
\label{perception}

Robust motion planning requires real-time dynamic perception of the environment for situational awareness. Dynamic perception enables robots to detect, track, and model moving obstacles in real time, providing their current obstacle states (i.e., positions, orientations, and velocities), future trajectory predictions, and uncertainty estimates. The role and requirements of dynamic perception vary across different planning methods. Sampling-based and search-based planners often rely on obstacle trajectories, collision states, or safe time intervals; MPC-based planners use predicted states and uncertainties to formulate safety constraints; and learning-based planners either rely on explicit agent states or learn them directly from raw sensor observations. Although this survey focuses on motion planning, it presents a review of dynamic perception techniques as a supporting component of motion planning.

This section briefly reviews representative perception techniques from a planning perspective. As shown in Fig.~\ref{fig:perception}, we categorize existing methods based on three sensing modalities: camera-based, LiDAR-based, and event-based. These modalities provide different information, such as semantic cues, geometric measurements, dynamic point filtering, object motion estimates, and low-latency event streams. Table~\ref{tab:dynamic_perception_comparison} summarizes the main strengths, limitations, and motion planning applications of the three sensing modalities.

\vspace{-10pt}
\subsection{Camera-based Methods}
\label{Camera_based_techniques}

Camera-based methods are widely adopted due to their high spatial resolution, rich semantic information, and relatively low hardware cost. Existing camera-based methods can be grouped into several categories. Since many recent systems integrate detection, tracking, mapping, prediction, and planning, the following grouping is based on each work's dominant perception method rather than all modules used in the system.

\vspace{3pt}
\subsubsection{Feature- and Flow-based Motion Detection} Camera-based dynamic perception methods often detect motion by analyzing temporal changes in visual features, calculating frame differences, or computing flow fields, usually with explicit compensation for camera ego-motion. Francis et al.~\cite{francis2015detection} used a Photonic Mixer Device (PMD) time-of-flight camera to acquire successive 3D range images and compute differential scene flow. By combining local--global flow estimation with Gradient Vector Flow (GVF)-based boundary detection, their method estimates the 3D motion and orientation of moving obstacles for path planning. Logoglu et al.~\cite{berker2017feature} proposed a feature-based moving-object detection pipeline for low-altitude aerial platforms. It compensates for ego-motion using feature matching and homography estimation, detects candidate moving regions by three-frame differencing, and filters parallax-induced false positives using geometric constraints.

\vspace{3pt}
\subsubsection{RGB-D People Detection and Tracking}

For pedestrian detection and tracking in human-centered dynamic environments, RGB-D sensors are often used to combine appearance, depth, and motion cues. Jafari et al.~\cite{jafari2014real} developed a real-time RGB-D people detection and tracking system for mobile robots and head-worn cameras. The system integrates the methods of visual odometry, region-of-interest (ROI) extraction, ground-plane estimation, pedestrian detection, and multi-hypothesis tracking. For producing dynamic obstacle information in crowded environments, it combines a) a fast depth-based upper-body detector for close-range pedestrians, and b) a GPU-based full-body Histogram of Oriented Gradients (HOG) detector for distant pedestrians.

\subsubsection{Low-Latency Onboard Perception}

Onboard perception must be computationally efficient and low-latency for high-speed obstacle avoidance. Oleynikova et al.~\cite{oleynikova2015reactive} designed a fully onboard stereo-vision system that produces Video Graphics Array (VGA)-sized disparity images using a Field-Programmable Gate Array (FPGA). Obstacles are extracted from U-disparity maps, approximated as ellipses in a short-term map, and used to generate feasible waypoints for planning. Lu et al.~\cite{lu2022perception} developed an onboard RGB-D perception system for quadrotors facing multiple small and fast-moving objects. For trajectory prediction, the method: a)  uses YOLO-Fastest~\cite{yolofastestv2} for RGB-based detection, b) extracts 3D positions from depth images, and c) introduces 3D-SORT to estimate object positions, velocities, and accelerations. Xu et al.~\cite{xu2024onboard} proposed a lightweight RGB-D dynamic object detection and tracking system for low-power robots. It combines depth-image- and point-cloud-based detectors to improve accuracy, associates obstacles across frames using point-cloud statistical features, and estimates their states with a constant-acceleration Kalman filter. It also optionally incorporates a light weight learning-based detector to extend the detection range and improve dynamic obstacle identification.

\begin{table*}[t]
\centering
\caption{Comparison of dynamic perception modalities.}\vspace{-3pt}
\label{tab:dynamic_perception_comparison}
\centering
\setlength\tabcolsep{5pt}
\begin{tabular}{l l l l}
\toprule
\specialrule{0.1em}{1pt}{1pt} 
\tabincell{l}{\textbf{Modality}} 
&\tabincell{l}{\textbf{Strengths}} 
&\tabincell{l}{\textbf{Limitations}} 
&\tabincell{l}{\textbf{Planning Applications}} \\
\toprule

\tabincell{l}{\textbf{Camera-based}} 
&\tabincell{l}{{High spatial resolution,}\\{rich semantic information,}\\{and low hardware cost}} 
&\tabincell{l}{{Sensitive to illumination changes}\\{and occlusion, limited FOV}} 
&\tabincell{l}{{Semantic navigation and}\\{socially aware navigation}} \\

\specialrule{0em}{2pt}{2pt}
\tabincell{l}{\textbf{LiDAR-based}} 
&\tabincell{l}{{Accurate geometric measurement,}\\{wide FOV, and robust to}\\{illumination changes}} 
&\tabincell{l}{{Limited semantic information and}\\{relatively high computational resources}} 
&\tabincell{l}{{Autonomous driving}} \\

\specialrule{0em}{2pt}{2pt}
\tabincell{l}{\textbf{Event-based}}
&\tabincell{l}{{Low-latency sensing for high-speed}\\{or challenging-illumination scenarios}} 
&\tabincell{l}{{Less informative in static or slowly changing}\\{scenarios and requires specialized algorithms}}
&\tabincell{l}{{Agile UAV navigation}} \\
\bottomrule
\specialrule{0.1em}{1pt}{1pt} 
\end{tabular}
\vspace{-1.0em}
\end{table*}

\vspace{3pt}
\subsubsection{Fusion, Tracking, and Mapping for Planning}

Recent systems increasingly combine geometric cues, visual detection, temporal tracking, and mapping to produce planning-ready obstacle states and environmental representations. Eppenberger et al.~\cite{eppenberger2020leveraging} proposed a stereo-camera-based system for real-time dynamic obstacle detection and tracking on computationally constrained robots. The method clusters noisy stereo point clouds to identify generic objects, classifies them as static or dynamic through inter-frame motion consistency, fuses the results with a visual people detector, and generates a 2D occupancy grid with estimated obstacle velocities for obstacle avoidance. Lin et al.~\cite{lin2020robust} presented an onboard vision-based detection and tracking module for micro aerial vehicle (MAV) navigation. Obstacles are detected from depth images and U-depth maps, represented as 3D boxes, and tracked with a Kalman filter to estimate their positions, velocities, sizes, and uncertainties. These estimates are then converted into ellipsoidal obstacle models for MPC-based planning.

Xu et al.~\cite{xu2023real} proposed a real-time RGB-D tracking and mapping system for UAV navigation in dynamic environments. The method generates obstacle region proposals from depth images, refines them with a static occupancy map, and uses Kalman and continuity filters to identify and track dynamic obstacles. It also removes dynamic-obstacle trails from the static map and predicts obstacle trajectories using a Markov chain model. Wang et al.~\cite{wang2021autonomous} developed an occlusion-aware dynamic perception module for onboard quadrotor navigation. Their method clusters depth-based point clouds, tracks obstacles with forward-propagating Kalman filters, and classifies them as static or dynamic through motion-based voting. Dynamic objects are represented as moving ellipsoids, while static points are fused into an occupancy grid map for trajectory generation. Xu et al.~\cite{xu2025intent} further addressed tracking loss caused by the limited onboard camera field of view (FOV) within an intent prediction-biased MPC framework. Their perception module combines point-cloud detection based on Density-Based Spatial Clustering of Applications with Noise (DBSCAN) 
algorithm~\cite{ester1996density}, U-depth detection, and YOLO verification, and propagating previously tracked obstacles with enlarged risk regions when they leave the camera view.

\vspace{3pt}
\subsubsection{Active Vision} Instead of passively relying on a fixed FOV, active vision methods adjust the sensing direction to improve the observability of dynamic obstacles and relevant free space. Chen et al.~\cite{chen2021active} proposed an active sense-and-avoid system in which a stereo camera is mounted on a servo-driven rotational mechanism to overcome the limited field-of-view of onboard vision. The sensing direction is planned to track dynamic obstacles, observe goal and flight directions, and explore unseen areas, providing perception support for uncertainty-aware collision checking and flight planning.

\subsection{LiDAR-based Methods}
\label{Lidar_based_techniques}

LiDAR-based perception provides accurate depth measurements and a wide FOV, making it particularly effective for detecting, tracking, and mapping dynamic obstacles in complex environments. Research in this area can be grouped into several representative  methods according to the dominant perception function of each method.

\vspace{3pt}
\subsubsection{Geometric Modeling, Segmentation, and Tracking}

These methods exploit the geometric structure of LiDAR measurements to estimate the states and shapes of dynamic objects and to separate the ground from obstacles. Asvadi et al.~\cite{asvadi20163d} proposed a 3D LiDAR perception framework for driving environments that combines multi-region piecewise ground-plane estimation with voxel-grid obstacle representation. After ground removal using the estimated planes, the method integrates consecutive scans with ego-motion information and applies discriminative analysis on voxel occupancy to distinguish static and moving obstacles.

Christie et al.~\cite{christie20163d} reconstructed moving vehicles by fusing sparse 3D laser scans with 2D image information. The method derives 3D correspondences from optical-flow matches, initializes registration with the 3-Point Random Sample Consensus (RANSAC) algorithm~\cite{fischler1981random}, and refines the alignment using the Iterative Closest Point (ICP) algorithm~\cite{chen1992object} with plane-normal cues. Kraemer et al.~\cite{kraemer2018lidar} represented dynamic object contours with adaptive 2D polylines instead of bounding boxes. The method jointly estimates object pose and shape, and uses free-space delimiters extracted from LiDAR scans to improve scan alignment and shape reconstruction.

\vspace{3pt}
\subsubsection{Multimodal Fusion}

Multimodal fusion methods combine the geometric accuracy of LiDAR with the semantic richness of cameras to improve 3D object detection and tracking. Ku et al.~\cite{ku2018joint} proposed a LiDAR-camera fusion framework that combines bird's-eye-view (BEV) LiDAR features with RGB image features for 3D object detection. The fused features are used to generate 3D proposals and refine the oriented bounding boxes with object categories in real time. Simon et al.~\cite{simon2019complexer} proposed Complexer-YOLO, a real-time 3D detection and tracking framework that fuses LiDAR point clouds with visual semantic segmentation. Semantic labels from RGB images are projected onto LiDAR points to form voxelized semantic point clouds, which are processed by an extended Complex-YOLO network to predict 3D bounding boxes. The framework further incorporates online multi-target tracking and introduces the Scale-Rotation-Translation score (SRTs) for efficient 3D box comparison.

\vspace{3pt}
\subsubsection{Dynamic Point Classification, Tracking, and Map Building}

Another line of work focuses on distinguishing dynamic and static LiDAR points 
%removing dynamic-object traces, 
and producing planning-ready static maps and dynamic object states. Yoon et al.~\cite{yoon2019mapless} proposed a mapless online method for point-level dynamic object detection in 3D LiDAR scans. The method compensates for motion distortion caused by spinning LiDAR sensors, compares recent scans using point-cloud discrepancies, and refines dynamic labels through motion-compensated free-space queries. Since it does not rely on prior maps, training data, or scene-specific assumptions, it can detect currently moving objects in a setting-independent manner.

Lim et al.~\cite{lim2021erasor} proposed a static 3D point cloud map building method that removes dynamic-object traces from accumulated LiDAR scans. It encodes query scans and map clouds with region-wise pseudo occupancy, uses a Scan Ratio Test to identify candidate dynamic bins, and applies Region-wise Ground Plane Fitting to preserve static ground points while rejecting dynamic residues. Fan et al.~\cite{fan2022dynamicfilter} proposed DynamicFilter, an online dynamic object removal framework for highly dynamic LiDAR environments. The framework consists of a scan-to-map front-end and a map-to-map back-end, integrating visibility-based and map-based strategies. The front-end performs fast visibility-based removal and map-based reverting to recover falsely removed static points, while the back-end estimates occupancy from multiple static submaps using visibility checks to approximate ray tracing.

Lu et al.~\cite{lu2024fapp} proposed a LiDAR-based framework for UAVs in dynamic cluttered environments. Its perception module clusters current point clouds using the DBSCAN algorithm~\cite{ester1996density} and compares them with recent historical point sets to classify clusters as moving, static, or unknown. The method tracks only dynamic clusters and estimates their states with a Kalman filter, while introducing covariance adaptation to handle dynamic objects with different motion patterns. After removing dynamic points, the remaining static points are used to build a local occupancy map for downstream obstacle avoidance.

\vspace{-6pt}
\subsection{Event-based Methods}
\label{Event_based_techniques}

Event-based vision is a bio-inspired sensing paradigm in which each pixel independently reports asynchronous events triggered by brightness changes rather than capturing full image frames at a fixed rate. This sensing mechanism has gained increasing attention for dynamic perception because it captures asynchronous brightness changes with microsecond-level latency. Compared with conventional frame-based cameras, event-based cameras are less susceptible to motion blur and can operate robustly in high-speed or low-light scenarios, making them well suited for fast dynamic obstacle perception.

Mitrokhin et al.~\cite{mitrokhin2018event} proposed an event-only method for moving object detection and tracking. It builds event-count images and time-images from short event windows, compensates global camera motion with a four-parameter model, and detects independently moving objects from residual motion inconsistencies before Kalman-filter tracking. Learning-based event perception has also been explored for dynamic obstacle avoidance. Sanket et al.~\cite{sanket2020evdodgenet} proposed a deep learning-based event-camera framework for quadrotor obstacle dodging. It uses shallow neural networks to deblur event frames, estimate ego-motion, and predict segmentation and optical flow of independently moving objects. Trained in simulation, it transfers to real-world onboard experiments without fine-tuning. 

To support low-power deployment, Ramesh et al.~\cite{ramesh2020low} proposed an event-based detection and categorization framework. It builds Rectangular Event Context Transform (RECT) features from local event activity, applies Principal Component Analysis (PCA) for compact representation, and uses a backtracking-free k-d tree for efficient matching. Its FPGA implementation enables real-time detection and classification. He et al.~\cite{he2021fast} further combined event and depth sensing to obtain planning-ready obstacle states. Their system compensates for ego-motion using IMU and depth data, segments moving regions from normalized mean-time images, and extracts object regions with iterative Gaussian fitting. It then tracks objects in the 2D image plane and estimates 3D trajectories by asynchronously fusing event and depth observations.

\vspace{-9pt}
\subsection{Summary of Dynamic Perception}
Dynamic perception provides obstacle-level information required by motion planners, including object states, velocities, predicted future trajectories, and uncertainty estimates. Camera-based perception methods offer rich semantic cues for human detection, scene understanding, and socially aware navigation. On the other hand, LiDAR-based methods provide accurate geometric measurements and wide-area depth perception, which are important for mapping and obstacle avoidance. Finally, event-based methods offer low-latency sensing for high-speed or challenging-illumination scenarios.

These modalities also have complementary limitations. Cameras are sensitive to illumination and occlusion, LiDAR provides limited semantic information and may require higher computational resources, and event-based sensors are less informative in static-dominant scenes. As summarized in Table~\ref{tab:dynamic_perception_comparison}, these trade-offs motivate multimodal perception systems that combine semantic detail, geometric accuracy, and temporal responsiveness. Such systems are especially important when prediction uncertainty, occlusion, crowd interaction, and real-time safety constraints affect planning decisions.
\section{Conclusion and Future Directions} \label{conclusions}

Motion planning in dynamic environments is a critical yet challenging problem in robotics, requiring robots to navigate safely and efficiently in the presence of moving obstacles. A wide range of methods has been developed to address this problem, spanning classical techniques, learning-based approaches, and hybrid strategies. This survey has reviewed 138 representative works, offering a structured taxonomy and detailed discussion across five major categories based on the concepts of sampling, graph search, model predictive control, learning, and other approaches, including velocity obstacles, artificial potential fields, and dynamic windows. We have highlighted the unique challenges of dynamic environments, including high-speed obstacles, prediction uncertainty, deadlock situations, and crowd--robot interactions, and examined how different approaches attempt to address them.

In addition, we emphasized the critical role of dynamic perception, which enables robots to detect, track, and predict the motion of surrounding agents and provides the essential input for safe and adaptive planning. Perception and planning are inherently interconnected: without reliable perception, even advanced planners cannot operate effectively in real-world dynamic settings. By positioning perception as a supporting module and planning as the central focus, this survey offers a holistic perspective on motion planning in dynamic environments and contributes to a deeper understanding of both the progress achieved and the challenges that remain.

Looking ahead, several open challenges deserve further attention as discussed below:
\begin{itemize}
\item First, learning-based planners still face a significant sim-to-real gap. Policies trained in simulation may fail under real-world scenarios due to noise, imperfect actuation, unmodeled agent behaviors, and rare safety-critical events. Improving robustness, transferability, and the ability to handle rare and unknown scenarios remains essential for practical deployment.

\item Second, verification and validation of complex data-driven planners remain difficult. Although learning-based planners provide strong adaptability, their safety properties are often harder to certify than those of classical planners. Future work should develop planning frameworks that combine the flexibility of learned models with formal safety mechanisms, such as reachability analysis, control barrier functions, chance constraints, and conformal prediction.

\item Third, scalability remains an open issue in extremely dense and dynamic crowds, where prediction uncertainty, multi-agent interaction, and computational constraints are tightly coupled. Efficient representations of crowd dynamics and scalable decision-making mechanisms are needed to support safe navigation in such scenarios.

\item Fourth, an important direction is the integration of perception and planning. Most existing methods still treat perception as a front-end module that provides obstacle states or predicted trajectories to the planner. However, robots operating in dynamic environments often need to actively decide where to look, which agents to track, and how much uncertainty can be tolerated during planning. This motivates tightly coupled active perception and planning frameworks, in which sensing actions and motion decisions are optimized jointly. Such integration is particularly important for occluded, cluttered, or human-centric environments, where incomplete perception can directly lead to unsafe or overly conservative behavior.

\item Finally, embodied AI techniques may further reshape motion planning in dynamic environments. GNNs provide structured representations for reasoning about multi-agent interactions, while semantic and language-guided planning offers a promising direction for incorporating high-level scene understanding, commonsense reasoning, and task-level decision support. Large language models and vision-language models are not the central focus of this survey, which primarily reviews algorithm-level motion planning methods. However, their increasing role in embodied agents suggests that future planners may integrate geometric reasoning, learned prediction, semantic scene understanding, and language-conditioned decision-making. Developing such integrated systems while maintaining real-time performance, interpretability, and safety guarantees remains an open problem.
\end{itemize}
\vspace{-0pt}

\balance
\bibliographystyle{ieeetr}
\bibliography{reference,dynamic_perception}

\end{document}